\documentclass[letterpaper, 10 pt, conference]{ieeeconf}  % Comment this line out if you need a4paper
\usepackage[colorlinks,bookmarksopen,bookmarksnumbered]{hyperref}

\usepackage{siunitx}

\usepackage{color}
\usepackage{amsmath}
\usepackage{amssymb}
\usepackage{prettyref}
\usepackage{graphicx}                      
\usepackage{booktabs}
\usepackage{multirow}
\usepackage{url}

\usepackage{subcaption}
\newrefformat{fig}{Figure~\ref{#1}}
\newrefformat{sec}{Section~\ref{#1}}
\newrefformat{tab}{Table~\ref{#1}}
\newrefformat{alg}{Algorithm~\ref{#1}}
\newrefformat{eq}{Equation~\ref{#1}}

\IEEEoverridecommandlockouts                              % This command is only needed if 
\title{\LARGE \bf CATCH-919 Hand: Design of a $9$-actuator $19$-DOF Anthropomorphic Robotic Hand}
\author{Zhong Zhang, Tao Han, Jia Pan$^\dagger$, and Zheng Wang%
\thanks{Z. Zhang, T. Han and J. Pan are with the Department of Mechanical and Biomedical Engineering, City University of Hong Kong. Z. Wang is with the Department of Mechanical Engineering, the University of Hong Kong. $^\dagger$ denotes the corresponding author. Email: jiapan@cityu.edu.hk}%
}

\begin{document}

\maketitle
\thispagestyle{empty}
\pagestyle{empty}

%%%%%%%%%%%%%%%%%%%%%%%%%%%%%%%%%%%%%%%%%%%%%%%%%%%%%%%%%%%%%%%%%%%%%%%%%%%%%%%%
\begin{abstract}
To achieve human-like dexterity for anthropomorphic robotic hands, it is essential to understand the biomechanics and control strategies of the human hand, in order to reduce the number of actuators being used without loosing hand flexibility.
To this end, in this article we propose a new interpretation about the working mechanism of the metacarpal (MCP) joint's extension and the underlying control strategies of the human hand, based on which we further propose a highly flexible finger design to achieve independent movements of interphalangeal (IP) joints and MCP joint. Besides, we consider the hyperextension of fingertip into our design which helps robotic finger present compliant and adaptive posture for touching and pinching. In addition, human thumb muscle functions are reconstructed in the proposed robotic hand design, by replacing 9 human muscle tendons with 3 cables in the proposed task-oriented design, realizing all 33 static and stable grasping postures. Videos are available at \url{https://sites.google.com/view/szwd}.
\end{abstract}

%%%%%%%%%%%%%%%%%%%%%%
% Outline
% 1. Introduction
% 	a) Self motion (definition, applications)
%   b) Approaches to self motion (random sampling, bnb)
%   c) Contribution in this paper
%
% 2. Related works
%   a) Problem formulation (rotation matrix)
%   b) BnB framework (contractor and bisector)
%
% 3. Algorithm
%
% 4. Experiments
%
% 5. Conclusion (future work)
%%%%%%%%%%%%%%%%%%%%%%

\section{Introduction}

The human hand has demonstrated tremendous dexterity in grasping and holding objects with various shapes. Many investigations suggest that such dexterity is highly related to the biomechanics of the human hand. Because of this reason, mimicking the biomechanical features of the human hand has long been considered as the gold standard for anthropomorphic hand designing in robotics. Developing such biomimetic hand is of significance to applications like prosthetics and industrial manufacturing, where the robotic hand is required to share similar, or even identical features as its human counterpart in the shape, structure, and function.

\begin{figure}[tb]
\centering
  \includegraphics[width=\linewidth]{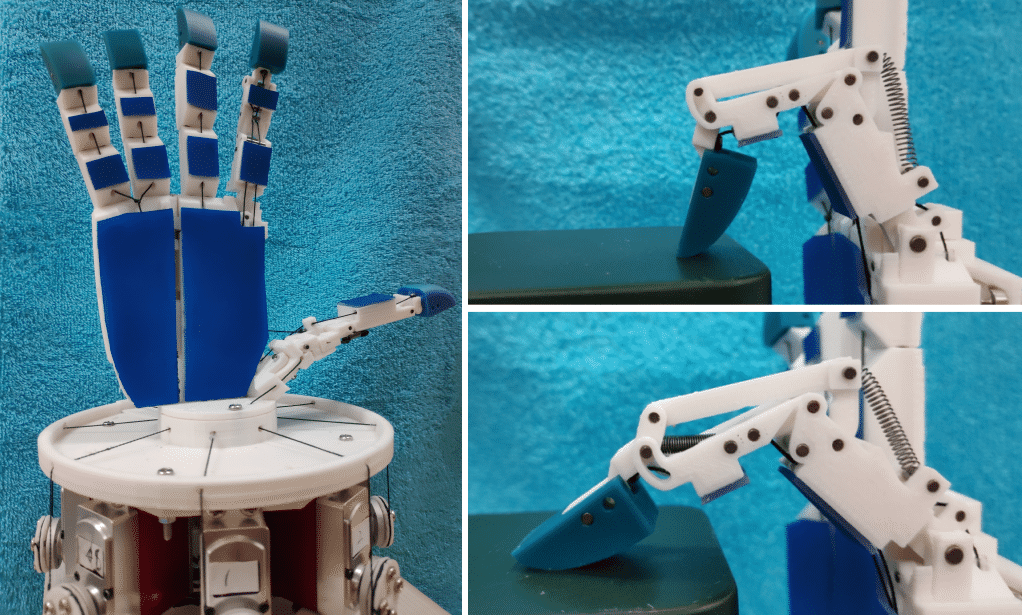}
  \caption{Our fully-assembled CATCH-919 Hand (the Cable-driven Anthropomorphic Tendon-Controlled Hand with 9 actuators and 19 DOFs). Left: the palmar aspect of the robotic hand. Top Right: an awkward posture of the index finger for touching. Bottom Right: a compliant posture of the index finger for touching.} 
  \label{fig:whole_hand}
\end{figure}

However, migrating the biomechanics of the human hand into its robotic replica is still demanding from the perspective of engineering. One research direction~\cite{mouri2002anthropomorphic} in the anthropomorphic hand field follows the design of classical robot arm by embedding one motor into each hand joint for separated rotation control with simplified mechanical structure. Approximating the kinematics of the human hand in such direct motor-driven manner can effectively decouple motions of different joints and brings convenience for control policy design. Unfortunately, this method also introduces some problems like low payload due to micro-actuators and high energy loss because distal joint's actuator becomes workload of proximal joint's actuator, which further reduces the entire system's load-bearing capacity.
% On one hand, the built-in motor needs be sufficiently micro to fit into the limited space in a hand joint, but a micro-actuator usually does not have enough load bearing capacity to support many hand grasping tasks. On the other hand, the motors placed in the fingertip joints will become the payload to other motors, which further decreases the load bearing capacity of the entire system.
As so far, the soft hand appeals to many researchers working on grasping and manipulation due to its inherent compliance of soft materials and the consequent low control complexity~\cite{deimel2016novel}. A soft hand can carry large payload compared its self-weight~\cite{zhou2017soft} and achieve all independent controls of joints~\cite{zhou2018bcl}. However, the soft hand may produce unwanted deformations due to its soft materials, and this will lead to imprecise control of in-hand manipulations. What's more, the actuators in the aforementioned soft hands are pneumatic devices which are too big to be integrated into humanoid robots. 

To solve above problems including low payload, the uncertainty of control and the large volume of actuators, we focus on designing cable-driven anthropomorphic hand in a fashion similar to human hand tendons. One primary benefit of employing such cable-driven design is that the actuators do not need to fit into the joint space, and thus they can be big enough to have sufficient load-bearing capacity and can be integrated into the robotic forearm like that of a human. However, compared with the motor-driven design, the cable-driven design usually requires many more actuators to achieve the same degrees of freedom (DOFs) in finger movement. In particular, the motor-driven design only needs one actuator to control the forward and backward rotations of a joint while the cable-driven method requires two actuators to control these two types of rotations separately.
% One typical example interpreting this phenomenon is that the motor-driven design only needs one actuator to control the forward and backward rotations of a joint, while the cable-driven method requires two actuators to control these two types of rotations separately.
In order to reduce the actuator number without loosing flexibility, many cable-driven anthropomorphic hand developed, including Shadow hand~\cite{hand2013shadow}, Yale openhand~\cite{ma2013modular} and other recent designs~\cite{wiste2017design,bhadugale2018anthropomorphically}. However, because most designers consider the actuator number as a priority in the mechanizing process, they discard some critical biomechanical features of the human hand, which undoubtedly leads to discrepancies between the real human hand and their proposed systems. Due to the lack of biomechanical features which are essential for human-like dexterity, these existing designs cannot achieve independent movements of every joint or they have to use more actuators compared with human tendons.

% Because of this reason, the anatomically corrected testbed (ACT) hand~\cite{wilkinson2003extensor} introduced a cable-driven design using four actuators to achieve all possible human finger postures with an artificial \textit{extensor mechanism}~\cite{extensor}. In order to explore potential biomechanical features and investigate neural control strategies of the human hand, they also studied the humam hand’s skeletal structure~\cite{Weghe:2004:ICRA}, kinematics of the thumb~\cite{Chang:2006:ACT} and variable moment arms for the index finger and thumb~\cite{Deshpande:2008:UVM,Deshpande:2010:TBE,niehues2017variable} to promote their hand more similar like human hand in anotamy.
To explore potential biomechanical features and investigate neural control strategies of the human hand, the anatomically corrected testbed (ACT) hand~\cite{Deshpande:2013:MAC} has developed through studying extensor mechanism~\cite{wilkinson2003extensor}, skeletal structure~\cite{Weghe:2004:ICRA}, kinematics of the thumb~\cite{Chang:2006:ACT}, variable moment arms for the index finger~\cite{Deshpande:2008:UVM,Deshpande:2010:TBE} and for the thumb~\cite{niehues2017variable} to promote the similarity of their hand to human hands in anatomy. 
However, its internal joints are still connected by hinges and gimbals, which are incorrect anatomically and prevent robotic fingers from achieving the human-level dexterity. In particular, the finger joints are stabilized by the dense irregular connective tissue that is able to deform elastically. Thus, we cannot simply regard joints as fixed hinges or gimbals, especially for the saddle joint and condyloid variety of the thumb~\cite{kapanji1983physiology}.
In order to preserve the joints' biomechanics for the hand dexterity, a highly biomimetic anthropomorphic hand~\cite{Xu:2016:DHB} replaced hinges and gimbals with artificial joint capsules, crocheted ligaments, and laser-cut extensor hood. However, their work focused on the mapping from cables to electromyography (EMG) signals and tele-operation, while how to explain the biomechanical advantages of the replicated joints remains a challenging  problem.

In addition, all existing cable-driven anthropomorphic hands whose control strategies mimic humans still suffer from several limitations:
\begin{itemize}
\item{\textbf{Joints cannot be controlled independently}}:
Cable-driven design is difficult to decouple motions of joints. Although ACT hand~\cite{wilkinson2003extensor} realized the function of independent control for metacarpal (MCP) joint, it still cannot control IP joints independently due to its incorrect anatomical position of interosseus~\cite{interossei} and obscure about the working mechanism of the MCP joint's extension.
% In other words, although robotic hands in previous work can achieve all static postures like human hand, it's still challenging for independent control of MCP joint and IP joints in dynamic movements.
\item{\textbf{Without considering finger postures under external forces}}:
Previous work did not consider the possible posture under the influence of external forces, and thus the existing
cable-driven systems cannot respond to external forces in a
compliant and adaptive manner as the human hand, which
will lead to awkward postures in tasks like touching (as shown in the top right of \prettyref{fig:whole_hand}).
\item{\textbf{Web-like extensor mechanism is complex to replicate}}: 
Extensor mechanism is a web-like collection of tendinous material that help distal phalangeal and proximal phalangeal joints flex and extend simultaneously. This structure is troublesome and time-consuming to replicate into the anthropomorphic hand's design by using the rubber design~\cite{Xu:2016:DHB} or crocheted design~\cite{xu2011design}.
\item{\textbf{Thumb cannot fit contact surfaces without pronation DOF}}:
In previous work, the thumb has 4 DOFs which are sufficient for thumb's fingertip to reach any point in 3-D plane. However, there is no additional DOF with rotational axis along the proximal or distal phalanx, but such DOF is essential for thumb to fit contact surfaces.
\end{itemize}

% In this work, we focus on designing a cable-driven anthropomorphic hand which can generate more realistic postures for practical applications. One of our main contributions is that the proposed robotic hand is of great dexterity, and can pass all challenging tests defined in the GRASP taxonomy~\cite{Feix:2016:GTH} by performing thirty-three standard grasping postures in different tasks.
In our work, we focus on designing a Cable-driven Anthropomorphic Tendon-Controlled Hand with 9 actuators and 19 DOFs (CATCH-919 hand) whose index finger can be controlled independently for IP and MCP joints and achieve all possible postures with the consideration of the influence of external forces on the fingertip. The resulting hand design can pass all challenging tests defined in the GRASP taxonomy~\cite{Feix:2016:GTH}. Our contributions can be summarized as follows:
\begin{itemize}
\item{\textbf{Accomplish independent control for index finger joints}}:
By understanding the working mechanism about MCP joint's extension according to latest anatomical study, we propose a new interpretation about control strategies of human index finger within sagittal plane, which can solve the problem of independent control of index finger joints. 
The same technique can be also applicable for the independent control of middle, ring and little finger joints, which however is not necessary for accomplishing the grasping tasks in GRASP taxonomy~\cite{Feix:2016:GTH} and thus is not implemented in our design.
\item{\textbf{Consider finger postures under external forces in design}}:
% We consider the \textit{hyperextension}~\cite{bworld} of the index fingertip under external forces into our robotic finger design, and thus
Our anthropomorphic hand can accomplish compliant posture for the touching task (as shown in the bottom right of \prettyref{fig:whole_hand}). Such improvement can effectively enlarge the contact area between the index fingertip and the touched object, which is crucial to make the fingertip sense more surface area and generate massive forces for the task.
\item{\textbf{A novel four-bar linkage to mimic extensor mechanism}}:
We design a novel four-bar linkage to replace IP joints with a similar biomechanical feature of extensor mechanism which helps DIP and PIP joints flex and extend simultaneously~\cite{landsmeer1949anatomy}. This structure has clear kinematics and it is convenient to fabricate and assemble.
\item{\textbf{5-DOF thumb with 3 actuators by task-oriented design}}:
In order to pass all challenging tests,
we dismantle and combine muscles' functions in the thumb and replace 9 human muscle tendons with 3 cables according to the task-oriented design. Our robotic thumb can use only 3 actuators to control 5 DOFs in steps while the thumb in the previous work~\cite{Xu:2016:DHB} only has 4 DOFs with the same number of actuators.
\end{itemize}
%%%%%%%%%%%%%%%%%%%%%%
% Related Work Outline
% 	a) interval arithmetic, bnb framework
%   b) problem formulation
%%%%%%%%%%%%%%%%%%%%%%

\section{Control strategies of human index finger}
\label{sec:related_work}

The controversy about working mechanism of MCP joint's extension in anatomy prevents researchers from figuring out control strategies of human hand. In this section, we will discuss two different mechanical structures of the index finger according to the classical anatomical theory and the latest medical research to propose a new interpretation about control strategies for achieving independent control of IP and MCP joints. We also interpret the muscle activation when fingertip is extended by external forces, which is not considered in the previous work about anthropomorphic hand.

\begin{figure*}
\centering
\begin{subfigure}[b]{0.24\textwidth}
\includegraphics[width=0.95\linewidth]{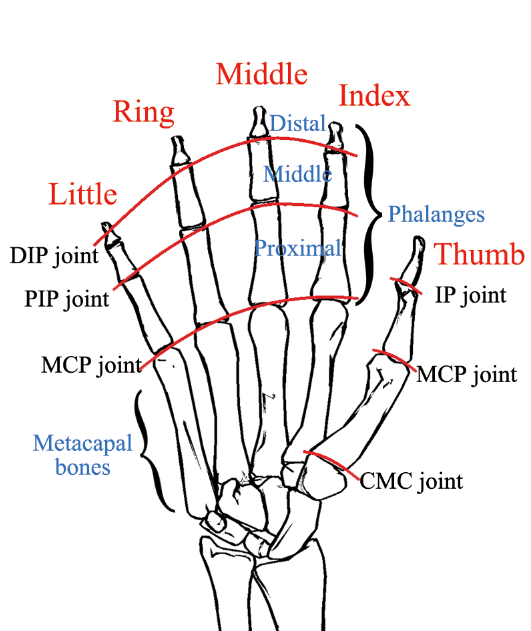}
\caption{}
\label{fig:whole_handbone}
\end{subfigure}
\begin{subfigure}[b]{0.24\textwidth}
\includegraphics[width=0.95\linewidth]{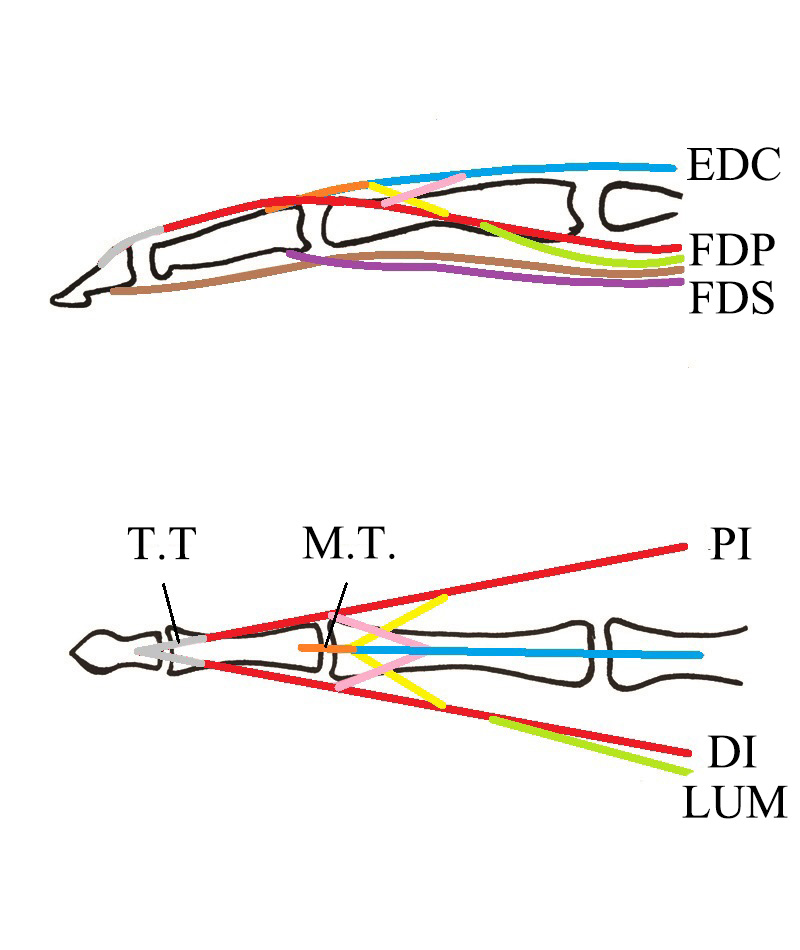}
\caption{}
\label{fig:index_tendon}
\end{subfigure}
\begin{subfigure}[b]{0.24\textwidth}
\includegraphics[width=0.95\linewidth]{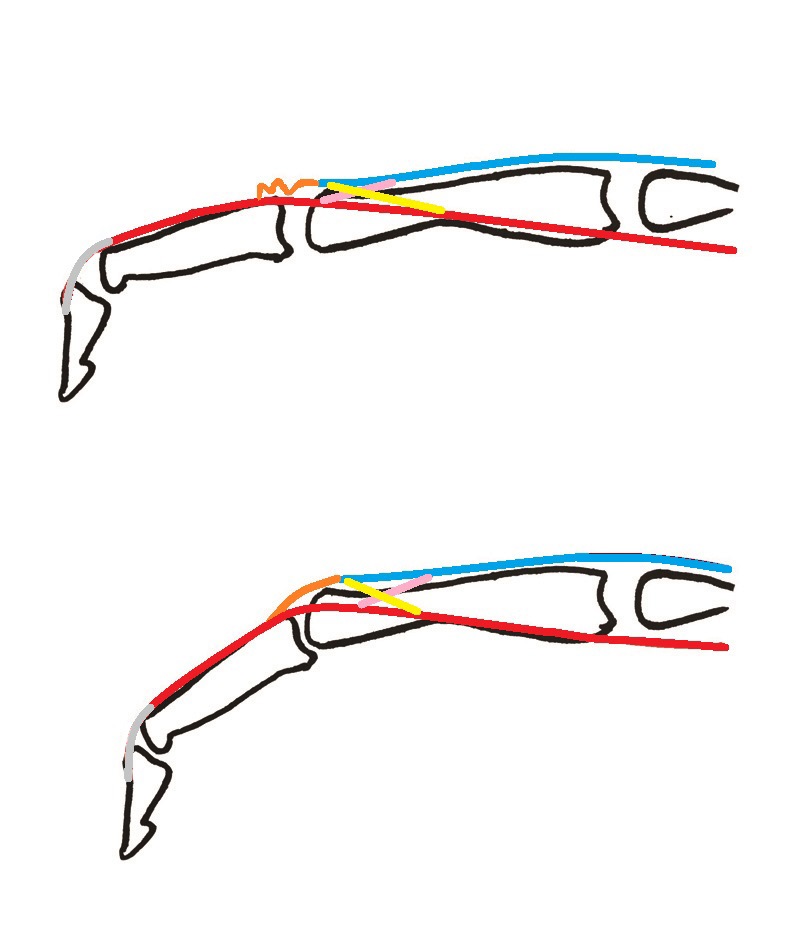}
\caption{}
\label{fig:extensor_mechanism_flex}
\end{subfigure}
\begin{subfigure}[b]{0.24\textwidth}
\includegraphics[width=0.95\linewidth]{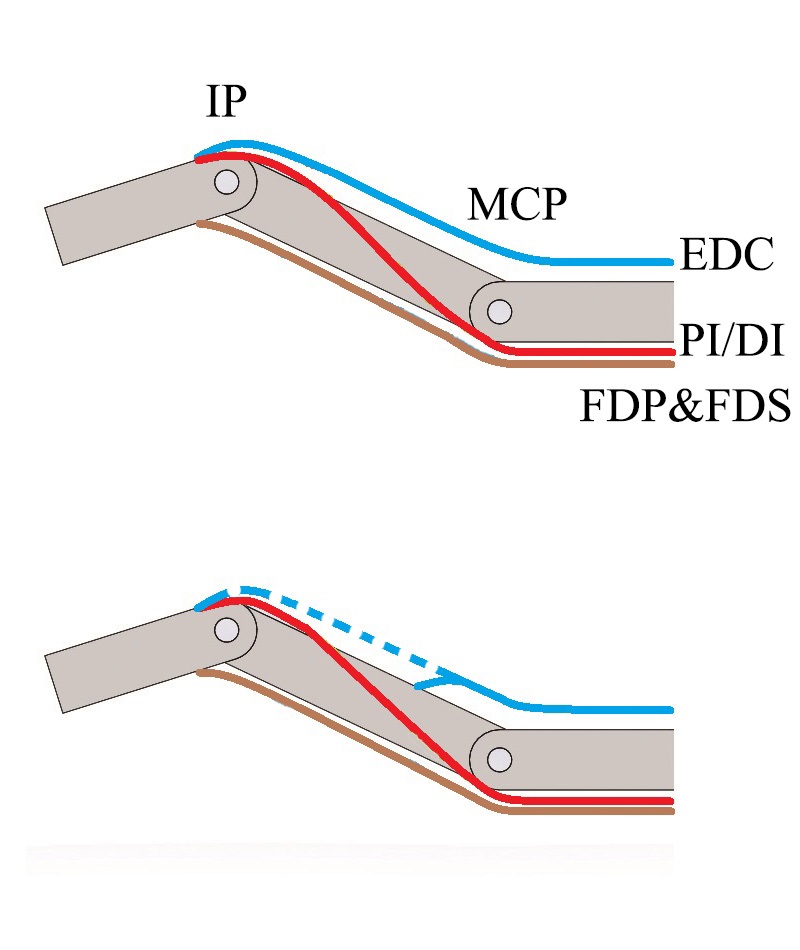}
\caption{}
\label{fig:MCP_2_model}
\end{subfigure}
\caption{Illustration of the finger structure. (a) shows the palmer view of human hand. (b) shows the distribution of tendon on the index finger. The top part is the lateral view of the index finger, where the blue line is EDC tendon, the brown line is FDP tendon, and the purple line is FDS tendon. The bottom part shows the the dorsal view of the index finger's extensor mechanism, where the red lines are lateral tendons connecting with PI and DI respectively, the green line is connected with LUM, the pink lines are lateral extensor bands, the yellow lines are medial extensor bands, the orange line is the medial tendon, and the gray lines are terminal tendons. Top of (c): DIP joint flexes when PIP joint is kept extended by external forces and EDC loses effective force on the middle phalanx due to the slack medial tendon. Bottom of (c): without external forces, the finger is kept on this state by FDP's flexion and retinacular ligament~\cite{landsmeer1949anatomy} which is not shown here. When EDC extends the finger, the above process repeats in the reverse order. (d) illustrates two mechanical structures of the index finger on the sagittal plane. In the top structure, EDC exerts effective forces on IP joints directly. In the bottom structure, EDC exerts effective forces on MCP joint directly via a line attached to the proximal phalanx. The dash line indicates the line that exists but is functionally unimportant in our design.} 
\label{fig:index}
\end{figure*}

\subsection{Index finger illustration}
As shown in \prettyref{fig:whole_handbone} and \prettyref{fig:index_tendon}, the index finger has three joints which are called distal interphalangeal (DIP) joint, proximal interphalangeal (PIP) joint, and metacarpal (MCP) joint respectively and it is controlled by 7 muscles: flexor digitorum profundus (FDP), flexor digitorum superficialis (FDS), lumbrical muscle (LUM), 2 interossei including palmar interosseous muscle (PI) and dorsal interosseous muscle (DI), extensor digitorum communis (EDC), and extensor indicis (EI). But EI does not exist in other fingers and it only strengths the independent ability to extend the index finger~\cite{extensor_indicis}. As a result, we omit this muscle in our following research.

Among these muscles, FDP is inserted in the palmar base of the distal phalanx and its primary function is to flex DIP, PIP and MCP joints. FDS is inserted in the palmar base of the middle phalanx and primarily a flexor of PIP and MCP joints. EDC is connected with an extensor mechanism that is inserted on dorsal sides of distal and middle phalanges, and its main role is to produce extension of MCP joint. It also helps to extend both PIP and DIP joints. However, the main extensors of these joints are the interossei (PI and DI) and lumbricals (LUM), which also help to prevent the hyperextension of MCP~\cite{palastanga2006anatomy}. Biomechanical studies have shown that interossei (PI and DI) is more essential for IP joints' extension and MCP joint's flexion than lumbrical (LUM)~\cite{schreuders1996strength,buford2005analysis}. And the full of muscle spindles in lumbricals suggests that their main function is for the proprioceptive perception of the fingers rather than for the motion control~\cite{wang2014biomechanical}.
Hence, the human index finger accomplishes the sufficient flexibility and dexterity of 4 DOFs (3 flexion/extension and 1 abduction/adduction) by using only 5 tendons (i.e., FDP, FDS, EDC, PI, DI, but without LUM) along with the extensor mechanism which helps DIP and PIP joints flex and extend simultaneously (as shown in \prettyref{fig:extensor_mechanism_flex}).

\subsection{Working mechanism of MCP joint's extension}
% As described before, the index finger can be controlled by 5 muscles with no insertion in the proximal phalanx, so
Till now, the MCP joint's extension principle remains to be a controversial topic in anatomy, and a set of different explanations have been proposed.
Some early work believed that the small deep slip of EDC tendon, which is inserted in the proximal phalanx, plays an important role in MCP joint's extension. However, anatomical and radiological studies have proven that such small deep slip is inconsistent and lax at all functional MCP joint positions and thus is functionally inessential for MCP joint's extension~\cite{jan1996insertion}. Another explanation widely accepted nowadays is that the MCP joint's extension is accomplished via the sagittal bands which, acting as a sling or lasso, attach the extensor tendon to the base of the proximal phalanx~\cite{baratz2005extensor}.
However, this statement rarely has any reference and is also challenged recently by~\cite{marshall2018mechanics}, which concluded that the torque passing from the dorsal side of the middle phalanx by EDC tendon's tightening is the main factor for the MCP joint extension.
The new structure proposed in~\cite{marshall2018mechanics} is also not accurate from the perspective of force analysis. To explain this, we illustrate the index finger movement within sagittal plane according to this work in the top of \prettyref{fig:MCP_2_model}. Note that here we regard PIP and DIP joints as one joint when studying the MCP joint movement, because both joints will flex simultaneously due to the extensor mechanism when flexors contract. However, when flexors contract (i.e., the brown line in the top of \prettyref{fig:MCP_2_model}), it is difficult to only rotate IP joint and use EDC to keep MCP joint stable from the perspective of force analysis. As a result, this structure cannot fully explain the independent movements of IP and MCP joints. ACT hand~\cite{wilkinson2003extensor} also used the same structure to interpret their robotic finger movement. However, their interossei does not always stay in the palmar side of MCP joint, which is not anatomically correct~\cite{interossei}. Although their cable control strategies can achieve all static postures like human hand, they still cannot solve the independent control of IP joints on the course of dynamic finger motion. 

From the aforementioned force analysis, we conclude that the main tension from EDC tendon should be exerted on proximal phalanx directly, and thus the independent control of IP and MCP joints can be achieved using structure shown in the bottom of \prettyref{fig:MCP_2_model}. Note that here we ignore the tension influence of the DIP and PIP joints through EDC tendon (the dash line in \prettyref{fig:MCP_2_model}), because such function is not significant. In the following part, we will use this new model to analyze the mapping from human hand's postures to muscle activation within sagittal plane.

% \begin{figure}
% \centering
% \begin{subfigure}[b]{0.49\linewidth}
% \includegraphics[width=.95\linewidth]{figs/MCPmodel_1.png}
% \caption{}
% \label{fig:MCPmodel_1}
% \end{subfigure}
% \begin{subfigure}[b]{0.49\linewidth}
% \includegraphics[width=.95\linewidth]{figs/MCPmodel_2.png}
% \caption{}
% \label{fig:MCPmodel_2}
% \end{subfigure}
% \caption{Illustration of the index finger model from lateral views. In (a) and (b): The red solid line is connected with EDC and red dotted line means that path is functionally insignificant in movements for EDC. The blue line represents flexors' tendon including FDP and FDS's. The yellow line is connected with PI/DI.The two rotational joints are IP joint and MCP joint respectively where the IP joint is combination of DIP and PIP joints.} 
% \label{fig:MCPmodel}
% \end{figure}

\subsection{Mapping from postures to muscle activation}
\begin{figure}
\centering
\begin{subfigure}[b]{0.24\linewidth}
\includegraphics[width=\linewidth,height=1.5cm]{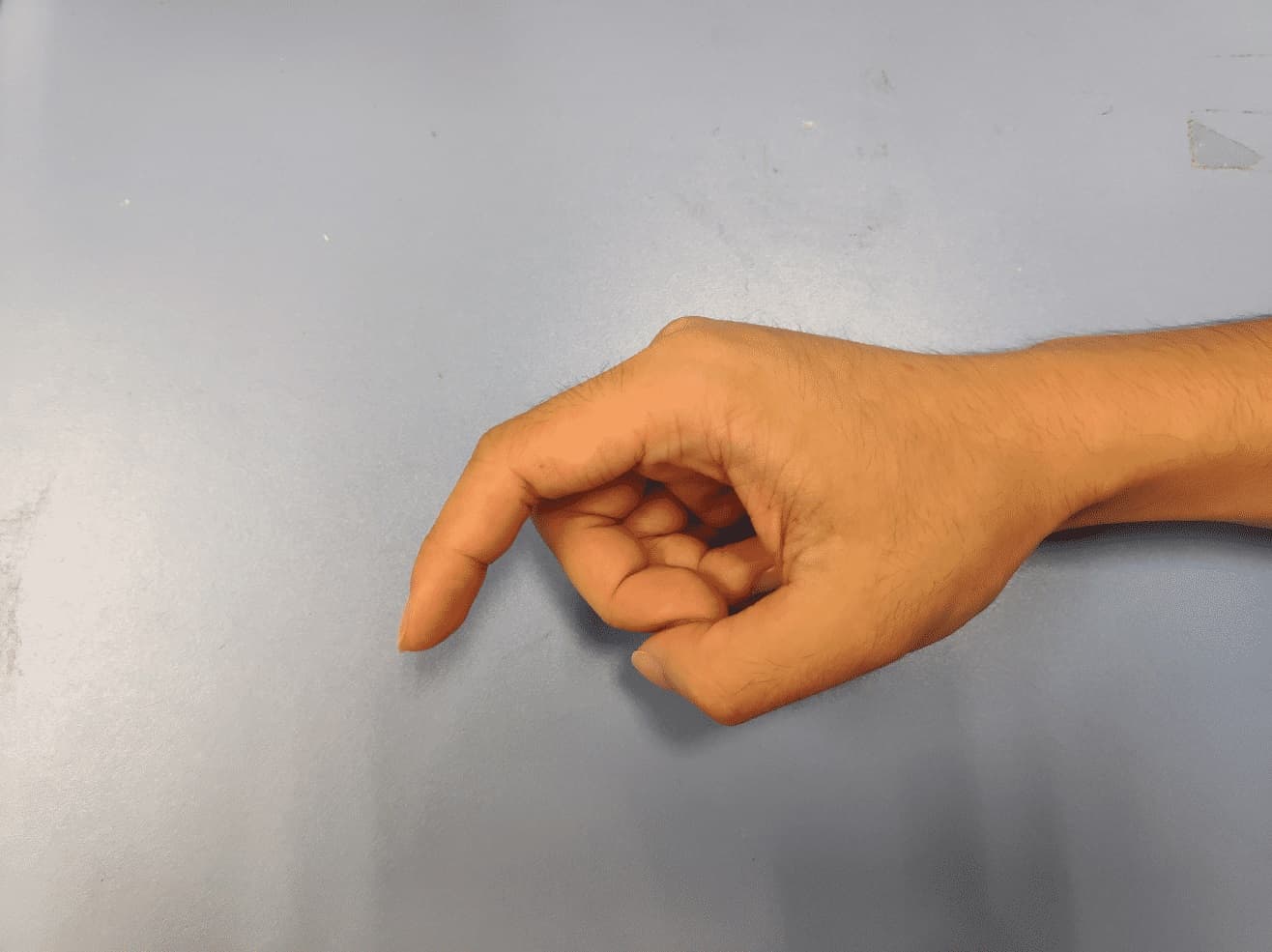}
\caption{}
\label{fig:a}
\end{subfigure}
\begin{subfigure}[b]{0.24\linewidth}
\includegraphics[width=\linewidth,height=1.5cm]{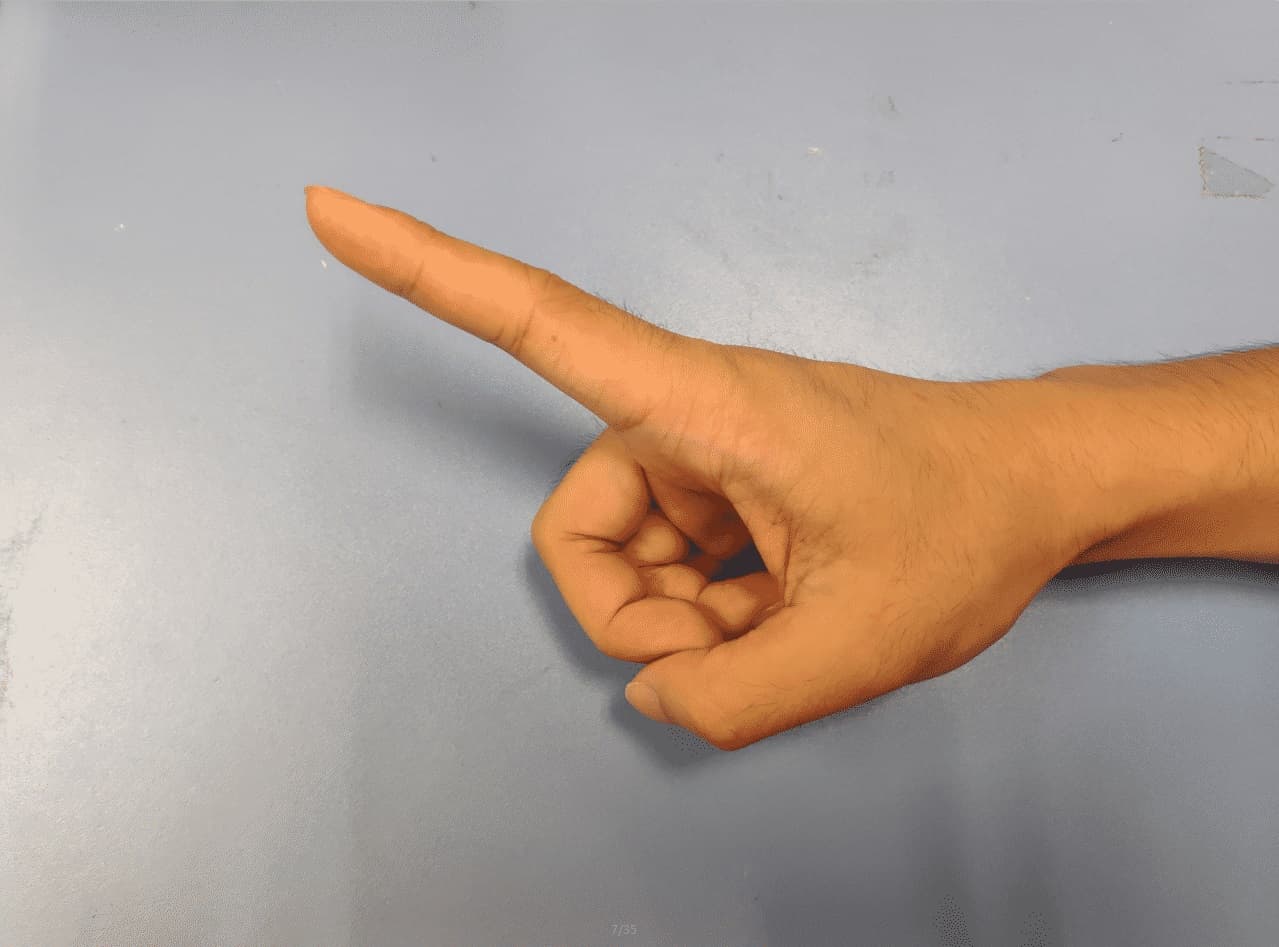}
\caption{}
\label{fig:b}
\end{subfigure}
\begin{subfigure}[b]{0.24\linewidth}
\includegraphics[width=\linewidth,height=1.5cm]{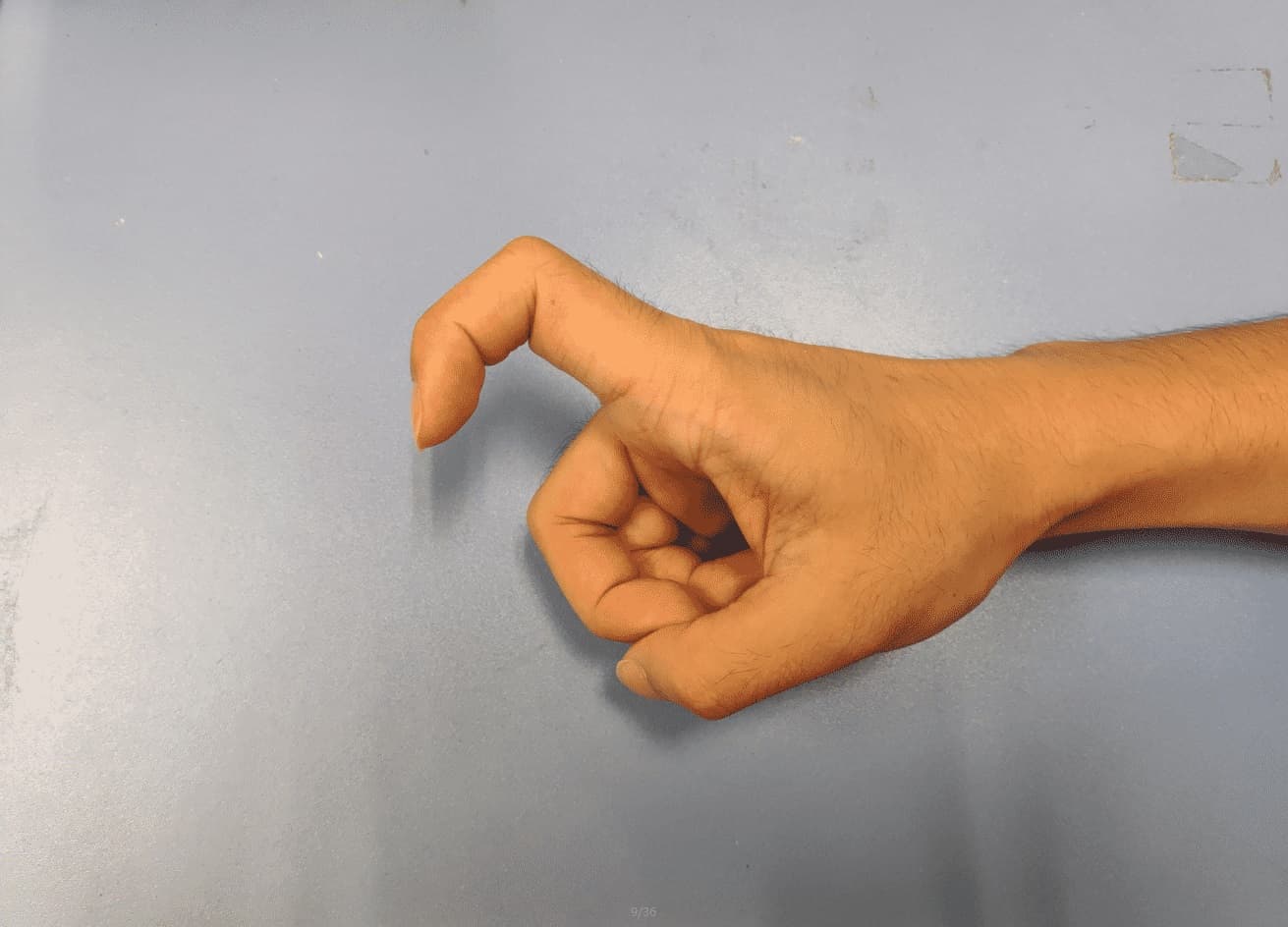}
\caption{}
\label{fig:c}
\end{subfigure}
\begin{subfigure}[b]{0.24\linewidth}
\includegraphics[width=\linewidth,height=1.5cm]{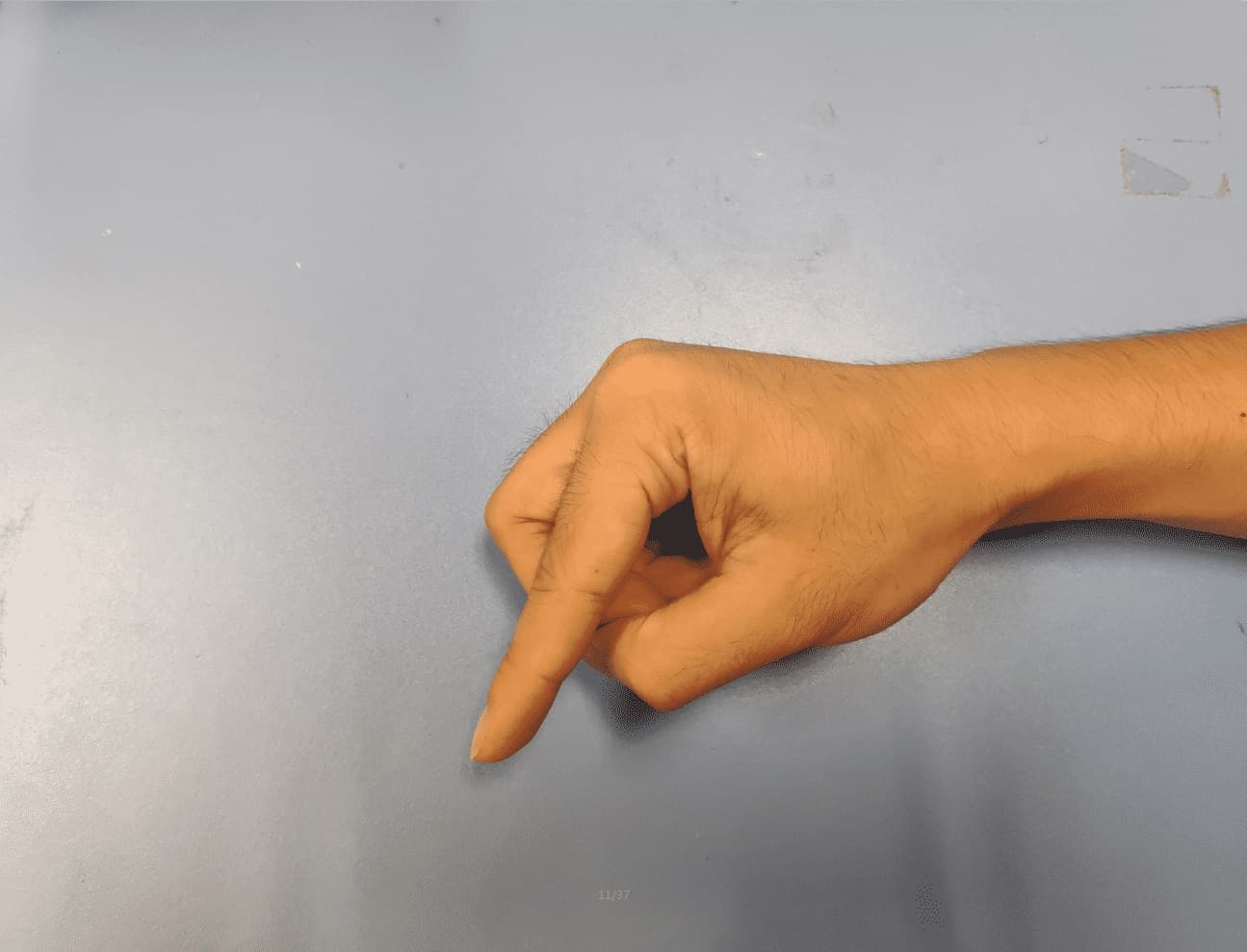}
\caption{}
\label{fig:d}
\end{subfigure}
\begin{subfigure}[b]{0.24\linewidth}
\includegraphics[width=\linewidth,height=1.5cm]{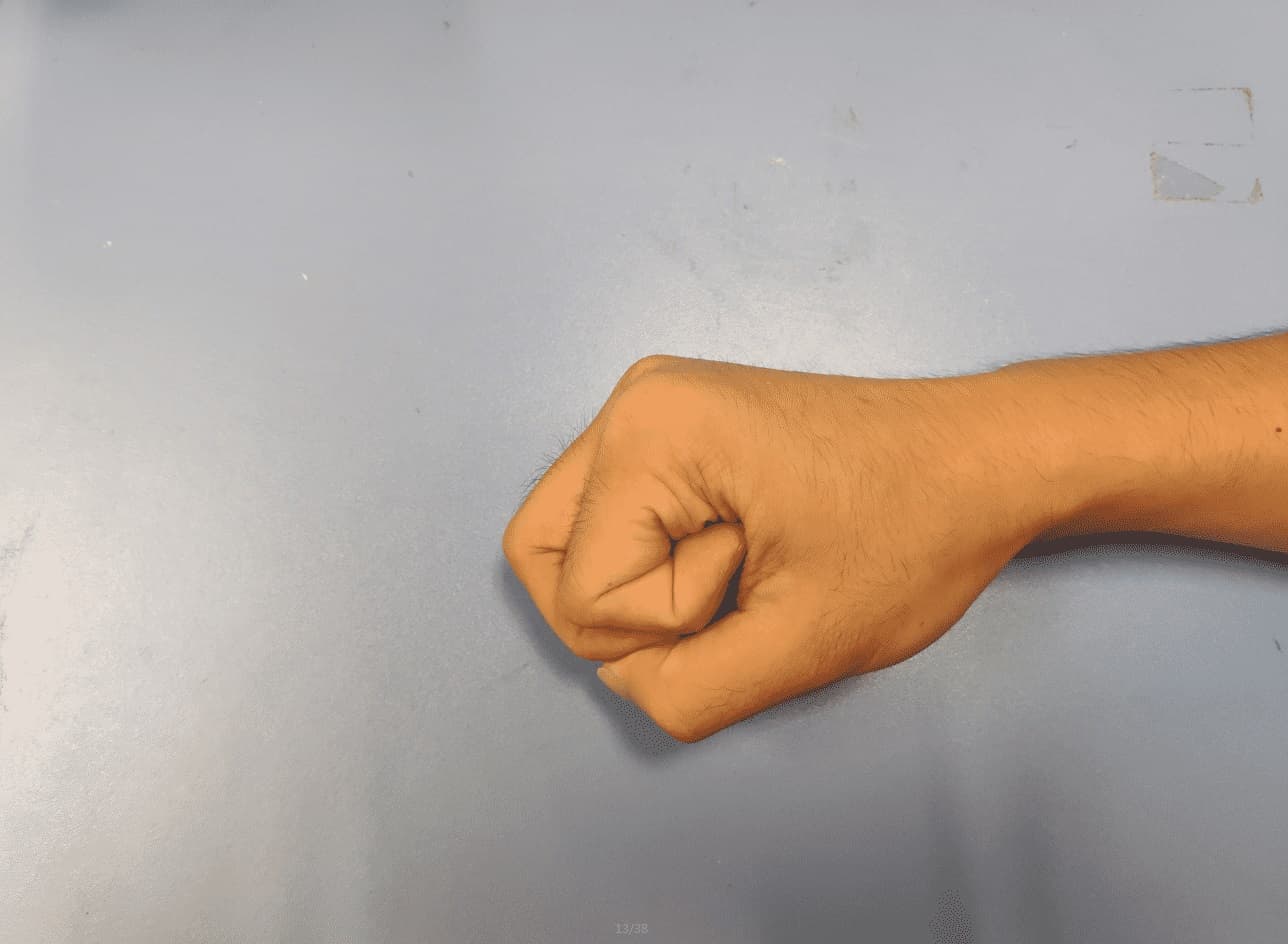}
\caption{}
\label{fig:e}
\end{subfigure}
\begin{subfigure}[b]{0.24\linewidth}
\includegraphics[width=\linewidth,height=1.5cm]{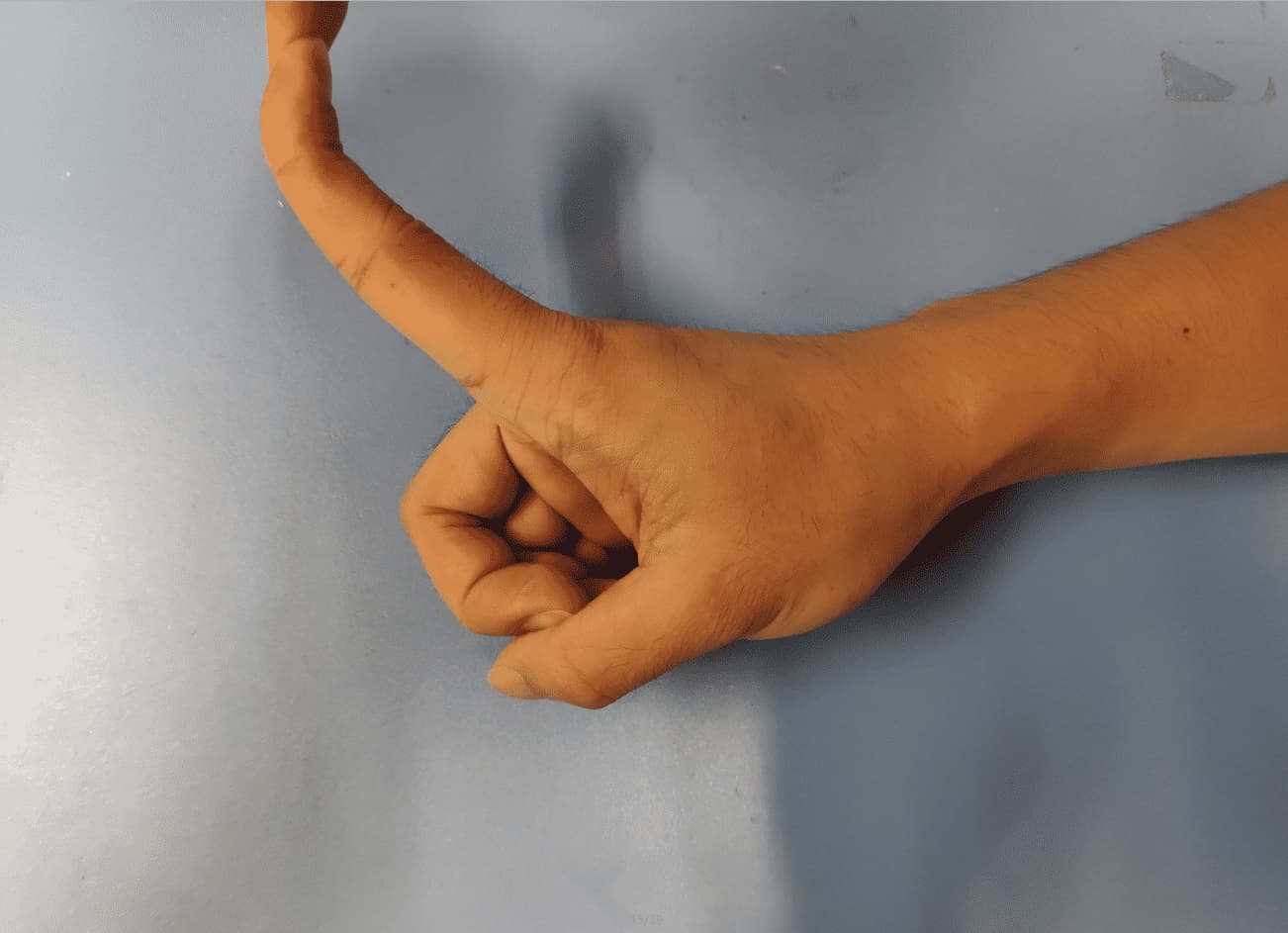}
\caption{}
\label{fig:f}
\end{subfigure}
\begin{subfigure}[b]{0.24\linewidth}
\includegraphics[width=\linewidth,height=1.5cm]{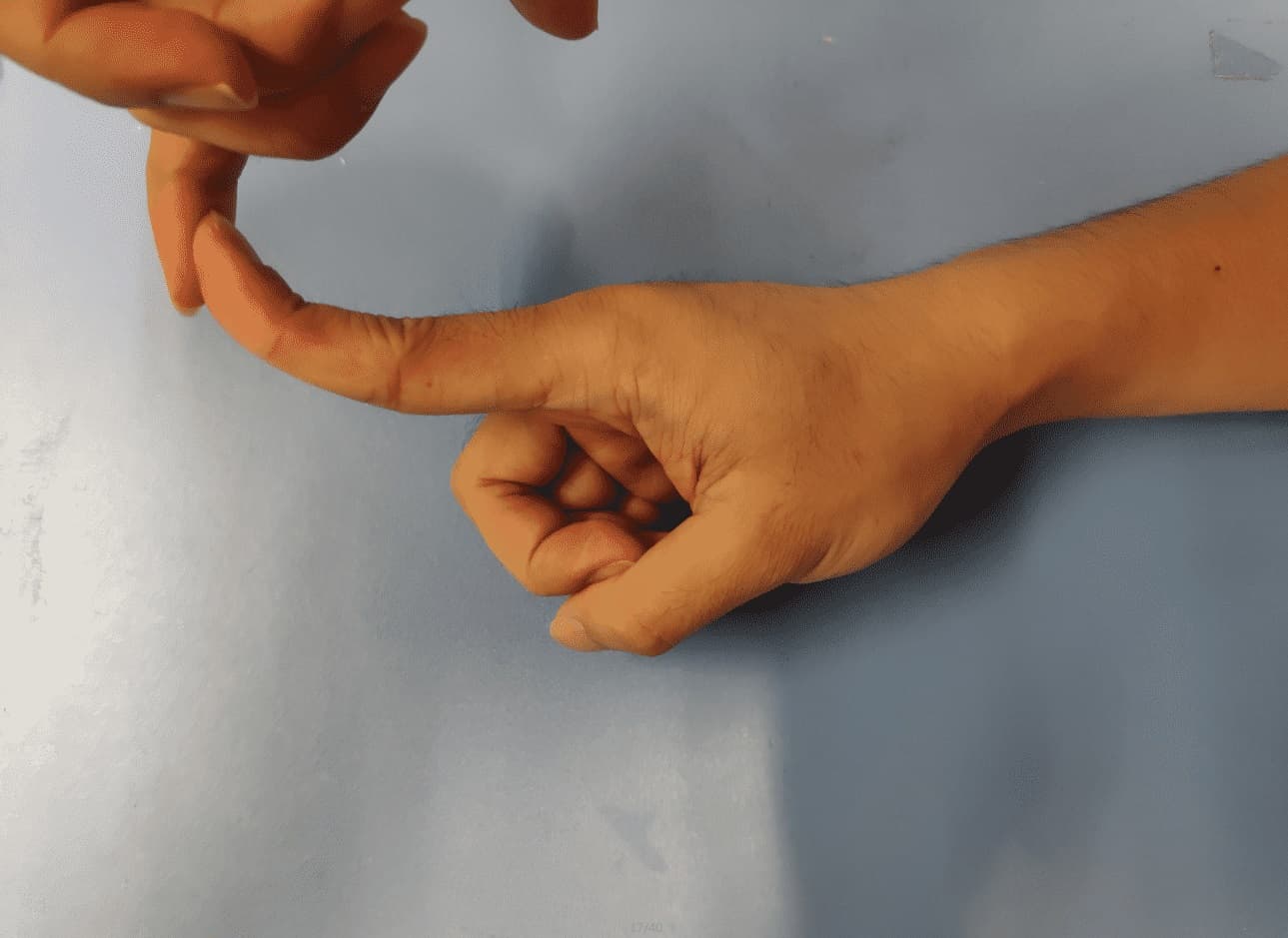}
\caption{}
\label{fig:g}
\end{subfigure}
\begin{subfigure}[b]{0.24\linewidth}
\includegraphics[width=\linewidth,height=1.5cm]{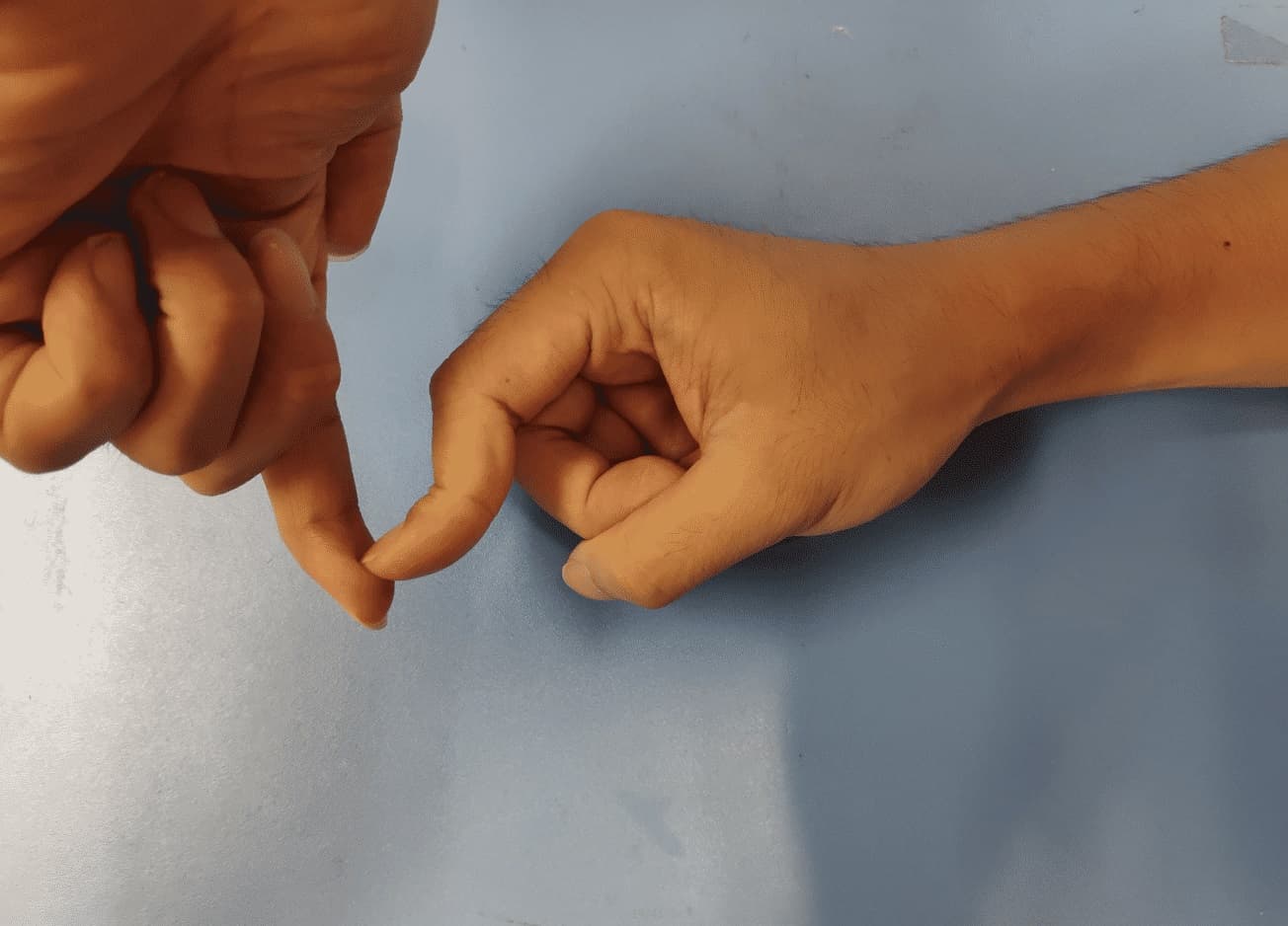}
\caption{}
\label{fig:h}
\end{subfigure}
\caption{Illustration of different hand postures. (a)-(e): finger postures without the influence of external forces. (f)-(h): finger postures forced by external forces on the palmar side of the fingertip. (a) and (f) are neutral positions where no muscle is activated.}
\label{fig:handposture}
\end{figure}

In sagittal plane, the finger postures without the influence of external forces can be categorized into 5 types as shown in \prettyref{fig:handposture}: (a) is neutral position where no muscle is activated. (b) represents MCP joint's \textit{hyperextension}\footnotemark ~or extension and IP joints' extension.\footnotetext{\textbf{hyperextension: } https://www.youtube.com/watch?v=NQWifmm\_qf4} (c) represents MCP joint's hyperextension or extension and IP joints' flexion. (d) represents MCP joint's flexion and IP joints' extension. (e) represents MCP joints' flexion and IP joints' flexion. The finger postures when the external force is exerted on the palmar side of fingertip and DIP joint stays in hyperextension state can be divided in 3 classes as shown in \prettyref{fig:handposture}: (f) represents the case when no muscle is activated. (g) represents PIP joint's extension no matter which MCP joint angle is. (h) represents PIP joint's flexion no matter which MCP joint angle is. Their mapping from postures within sagittal plane to muscle activation can be expressed in ~\prettyref{tab:human_mapping} where $+$ indicates that the corresponding muscle is activated. FDP's $(+)$ means that this muscle can be activated for resisting large external forces and EDC's $(+)$ indicates that this muscle can be activated to stabilize the MCP joint at any desired degree. 

\begin{table*}[!htbp]
\caption{Mapping from postures to muscle activation}
\label{tab:human_mapping}
\begin{center}
% \begin{tabular}{lcccc}
% \toprule
% Without external forces                 & FDP & FDS & PI/DI & EDC \\
% \midrule
% MCP (hyper)extension and IP extension &     &     & +     & +   \\
% MCP (hyper)extension and IP flexion   & +   & +   &       & +   \\
% MCP flexion and IP extension   & +   & +   & +     & (+) \\
% MCP flexion and IP extension   & +   & +   & +     & (+) \\
% \toprule
% Under external forces               & FDP & FDS & PI/DI & EDC \\
% \midrule
% PIP extension and any MCP angle& (+) & +   & +     &     \\
% PIP flexion and any MCP angle  & (+) & +   &       &     \\
% \bottomrule
% \end{tabular}
\begin{tabular}{|c|l|l|c|c|c|c|}
\hline
\multirow{2}{*}{\textbf{Posture}} & \multicolumn{2}{c|}{\multirow{2}{*}{\textbf{Description}}}                                       & \multicolumn{4}{c|}{\textbf{Muscle}}                                                          \\ \cline{4-7} 
                                  & \multicolumn{2}{c|}{}                                                                            & FDP                   & FDS                   & PI/DI                 & EDC                   \\ \hline
\prettyref{fig:b}                     & \multirow{4}{*}{without external forces} & MCP joint's (hyper)extension and IP joints' extension & \multicolumn{1}{l|}{} & \multicolumn{1}{l|}{} & +                     & +                     \\ \cline{1-1} \cline{3-7} 
\prettyref{fig:c}                  &                                          & MCP joint's (hyper)extension and IP joints' flexion   & +                     & +                     & \multicolumn{1}{l|}{} & +                     \\ \cline{1-1} \cline{3-7} 
\prettyref{fig:d}             &                                          & MCP joint's flexion and IP joints' extension          & +                     & +                     & +                     & (+)                   \\ \cline{1-1} \cline{3-7} 
\prettyref{fig:e}              &                                          & MCP joint's flexion and IP joints' flexion            & +                     & +                     &                      & (+)                   \\ \hline
\prettyref{fig:g}             & \multirow{2}{*}{under external forces}   & PIP joint's extension and any MCP joint's angle       & (+)                   & +                     & +                     & \multicolumn{1}{l|}{} \\ \cline{1-1} \cline{3-7} 
\prettyref{fig:h}           &                                          & PIP joint's flexion and any MCP joint's angle         & (+)                   & +                     & \multicolumn{1}{l|}{} & \multicolumn{1}{l|}{} \\ \hline
\end{tabular}
\end{center}
\end{table*}
\section{Development of our Robotic Hand}
\label{sec:development}

\begin{figure}
\centering
\begin{subfigure}{0.49\linewidth}
\includegraphics[width=\linewidth]{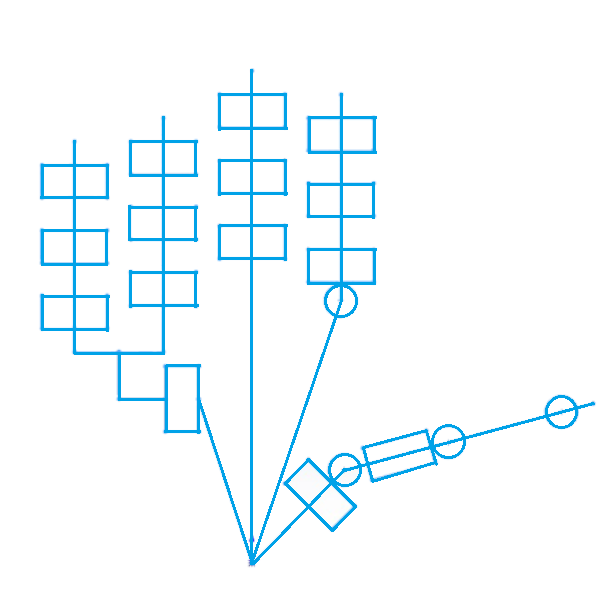}
\caption{}
\label{fig:simplified_structure}
\end{subfigure}
\begin{subfigure}{0.49\linewidth}
\includegraphics[width=\linewidth]{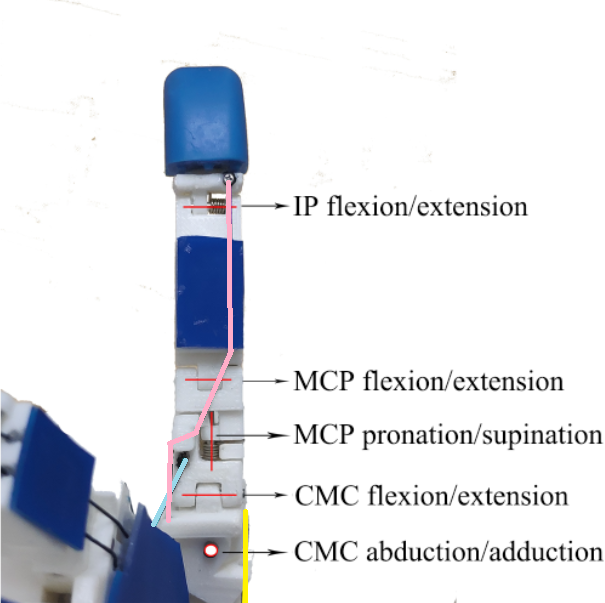}
\caption{}
\label{fig:thumb_topview}
\end{subfigure}
\caption{Illustration of our robotic hand model and robotic thumb. (a): the simplified 19-DOF robotic hand in the palmar view. (b): top view of robotic thumb.} 
\label{fig:simplified_handview}
\end{figure}

As shown in \prettyref{fig:simplified_structure}, in our design, every finger has 3 DOFs for flexion/extension except that the index finger has one more DOF to abduct/adduct, thumb has two more DOFs for abduction/adduction and pronation respectively, and the palm has an underactuated DOF that relies on ring and little fingers' flexion. In total, 9 Feetech servos\footnotemark ~(SCS40, \SI{40}{\kg\cm}) are used for controlling the finger movements: 1 for ring and little fingers, 1 for the middle finger, 4 for the index finger and 3 for the thumb.\footnotetext{\textbf{Feetech servo: } http://www.feetechrc.com/}

\subsection{Design of index IP joints}
\begin{figure}
\centering
\begin{subfigure}[b]{0.49\linewidth}
\includegraphics[width=\linewidth]{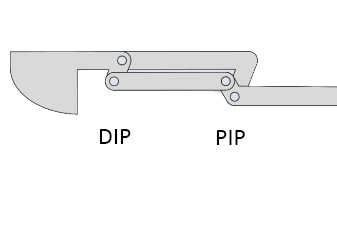}
\caption{}
\label{fig:fourlink_extension}
\end{subfigure}
\begin{subfigure}[b]{0.49\linewidth}
\includegraphics[width=\linewidth]{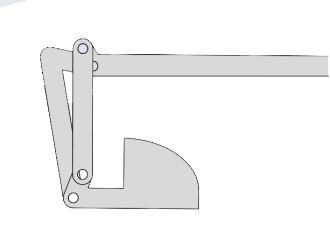}
\caption{}
\label{fig:fourlink_flexion}
\end{subfigure}
\begin{subfigure}[b]{1\linewidth}
\includegraphics[trim=0 60 0 10, clip, width=\linewidth]{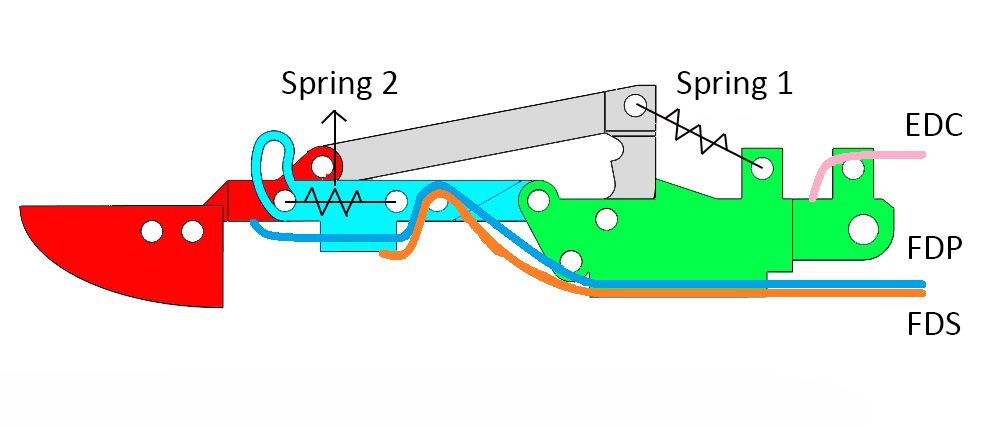}
\caption{}
\label{fig:index_whole_structure}
\end{subfigure}
\caption{Illustration of the four-bar linkage's movement and robotic index structure in the lateral view. (a) and (b): illustration of movements about four-bar linkage. (c): the lateral view of robotic index finger.} 
\label{fig:index_model}
\end{figure}

% According to the explanation from the previous subsection,
We replace the IP joints in the human hand by a novel four-bar linkage, which mimics the biomechanical feature of extensor mechanism in terms of enabling DIP and PIP joints flex and extend simultaneously. The four-bar linkage structure is illustrated in \prettyref{fig:index_model}.
Note that the simultaneous flexion and extension of DIP and PIP is the only biomechanical advantage of extensor mechanism with sufficient anatomical support. Other biomechanical advantages such as the gliding mechanism introduced in~\cite{Xu:2016:DHB} are not supported by sufficient evidence from the anatomical perspective~\cite{landsmeer1949anatomy}. Thus, according to our knowledge, our design does not sacrifice the biomechanical features that have been verified till now.

% Two rotational joints in \prettyref{fig:fourlink_extension} represent DIP and PIP joints respectively. Both these joints' rotational ranges are from 0 to 90 degrees.
To increase index finger's flexibility, we also consider how to realize the passive hyperextension of the DIP joint. This is an problem rarely touched in the cable-driven anthropomorphic hand community, but it is very important for pinching and touching tasks. In our design, we only consider the hyperextension of the DIP joint, because even though the human finger's PIP joint has hyperextension, its angle is very small compared with DIP joint. In addition, this design can also provide a more clear kinematics.

The whole design of IP joints is shown in \prettyref{fig:index_whole_structure}. The red region is the distal phalanx, the light blue region is the middle phalanx, and the green region is the proximal phalanx. Spring 1 implements the interossei's passive musculotendinous resistance to maintain PIP joint at 0 degree. Spring 2 connects distal and middle phalanges and implements FDP's passive musculotendinous resistance to maintain DIP joint at 0 degree. If fingertip's palmar side is carrying out an external force, it can extend along sliding chute, which is located in the distal side of light blue area.

% \begin{subfigure}[b]{0.45\textwidth}
% \includegraphics[width=.9\linewidth]{fig/100start}
% \caption{$100$ robots (in red) are moving to their goal positions (in yellow).}
% \label{fig:100_start}
% \end{subfigure}

\subsection{Design of the index MCP joint}
Although interossei can control the finger's abduction and adduction, it also influences the flexion of MCP joint and extension of IP joints. If we still use cable-driven mechanism in our anthropomorphic hand for abduction/adduction, the precise control of the finger would be difficult, because the lack of somatic sensation feedback as the human skin~\cite{bear2007neuroscience}, which may result in unstable control.
Hence, we use motor-driven design in the MCP joint for adduction/abduction directly and use the spring to implement the interossei's influence within sagittal plane, as shown in \prettyref{fig:index_whole_structure}.

\subsection{Design of the thumb finger}
From the geometric perspective, if a thumb wants to reach anywhere within a ring in the 2-D plane, it needs 2 joints; if it wants to change its terminal to any angle, it needs 3 joints in the 2-D plane; if it wants to reach anywhere in the 3-D plane on the basis of 3 joints before, it need one extra joint that is orthogonal to first three joints. But in order to better fit surfaces of other fingertips, the thumb needs a fifth degree of freedom that is orthogonal to other four joints. Thus, it is necessary to have at least 5 DOFs to ensure sufficient flexibilities for the thumb: 3 flexion/extension DOFs, 1 abduction/adduction DOF, and 1 pronation/supination DOF. 

The human thumb has three joints as shown in \prettyref{fig:whole_handbone}. Among these joints, the carpometacarpal (CMC) joint is commonly explained as a saddle joint~\cite{kapanji1983physiology} which is responsible for 1 flexion/extension DOF and 1 abduction/adduction DOF.
% Although the CMC joint requires not only saddle-shaped surfaces but also curved rotation axis that support rotation, sliding, translation, and pivoting motions~\cite{crisco2015vivo}(This is copied), its movement is so small compared with the movement in the flexion/extension or abduction/adduction direction.
Although the CMC joint has curved axis that allows rotation, sliding, translation, and pivoting motions~\cite{crisco2015vivo}, we discard this biomechanical feature because its precise locations of joint axes are still controversial.
The MCP joint of the thumb is a condyloid variety. Like every condyloid joint, it has two DOFs for flexion/extension and abduction/adduction. However, as a result of its complex biomechanics, it has a third DOF (pronation/supination) allowing axial rotation of the proximal phalanx, which is essential for thumb opposition~\cite{kapanji1983physiology}. To accomplish simpler kinematics, better control and more convenient assembly, we treat the CMC joint as a fixed 2-DOF universal joint for flexion/extension and abduction/adduction. For the MCP joint, we only reserve 1 flexion/extension DOF implemented using the hinge and 1 pronation/supination DOF implemented with the help of a small torsional spring, as shown in \prettyref{fig:thumb_topview}.

The human thumb has 9 motor muscles to achieve its great mobility and dexterity.
% and this abundance of dedicated muscles, as compared with the other fingers, determines its greater mobility and its essentiality~\cite{kapanji1983physiology}.
% But for our robotic hand, the thumb's basic tasks are assisting other four fingers to complete 33 static grasping postures.
However, it is unnecessary for our robotic thumb to assemble the same number of actuators according to the task-oriented design because we do not need the robotic thumb to accomplish very delicate movements.
To reduce the actuator number, we use springs to keep robotic thumb in extension state in order to save 3 extensors, including abductor pollicis longus, extensor pollicis brevis, and extensor pollicis longus.
Then we reconstruct the remaining 6 muscles' functions in the thumb and design 3 cables' layout as shown in \prettyref{fig:thumb_topview}. Among these wires, the yellow line is responsible for CMC joint's abduction. The light blue line is responsible for CMC joint's adduction and then CMC joint's flexion if CMC joint is fixed in the adduction/abduction direction. The pink line is responsible for MCP joint's flexion first. If the proximal or distal phalanx receives resistance from objects or MCP joint has reached the flexion limit, the pink line will be tightened to make MCP joint perform pronation movements
for better contact with the object surface and eventually drive IP joint flex.

\section{Performance of Our Robotic Hand}
\label{sec:experiments}

\begin{figure}
\centering
\begin{subfigure}{0.3\linewidth}
\includegraphics[width=\linewidth]{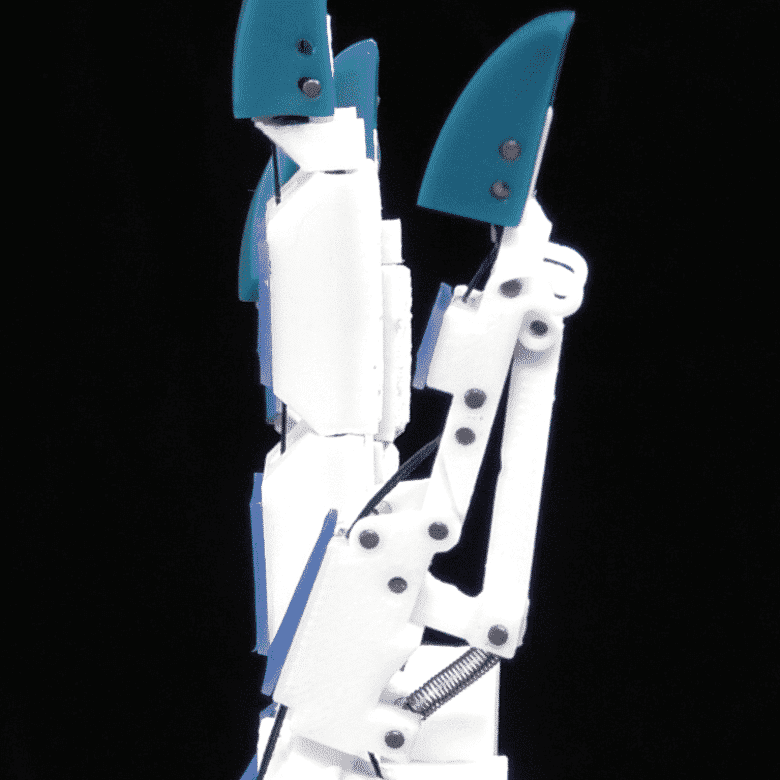}
\caption{}
\label{fig:aa}
\end{subfigure}
\begin{subfigure}{0.3\linewidth}
\includegraphics[width=\linewidth]{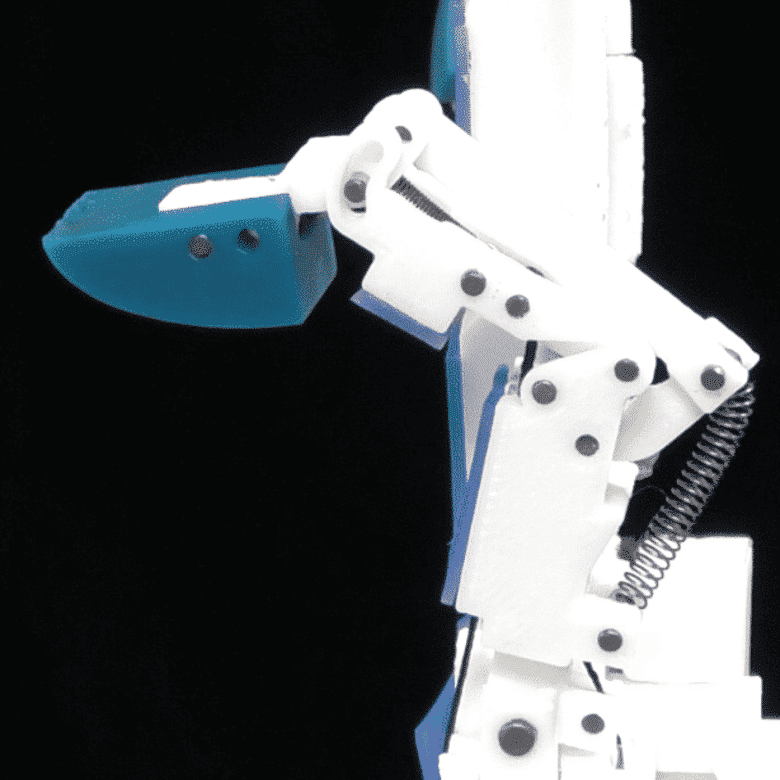}
\caption{}
\label{fig:bb}
\end{subfigure}
\begin{subfigure}{0.3\linewidth}
\includegraphics[width=\linewidth]{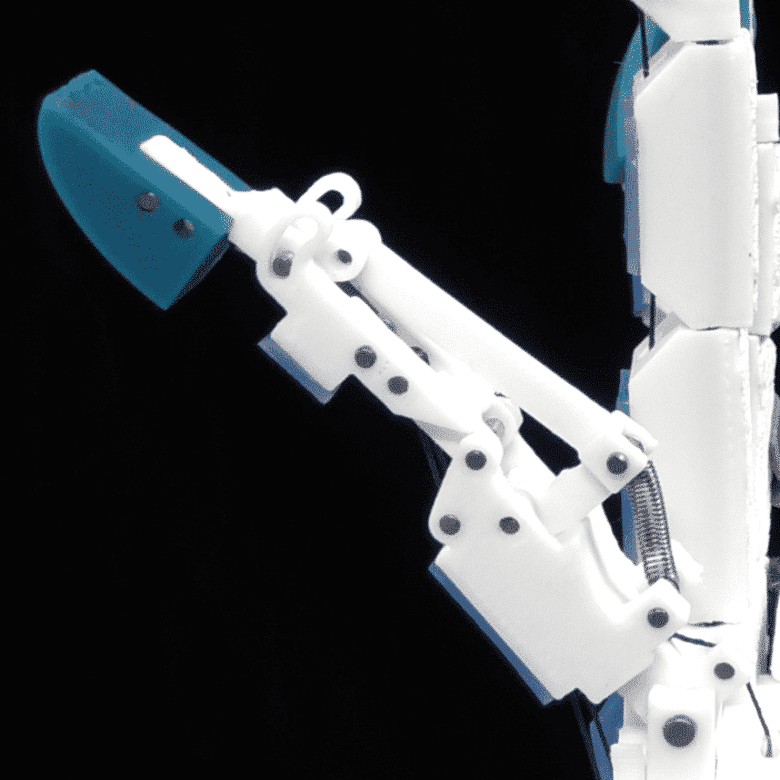}
\caption{}
\label{fig:cc}
\end{subfigure}
\begin{subfigure}{0.3\linewidth}
\includegraphics[width=\linewidth]{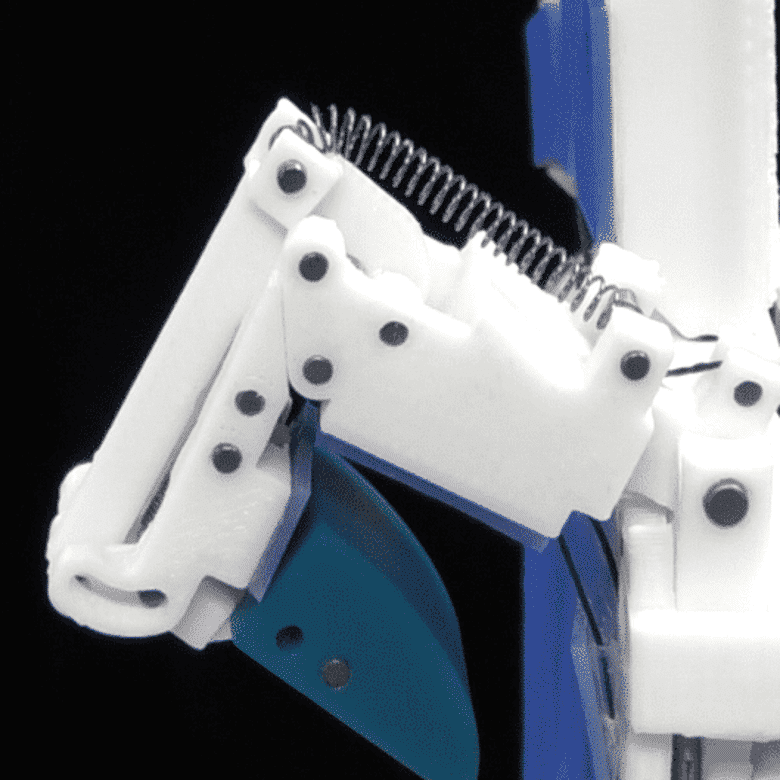}
\caption{}
\label{fig:dd}
\end{subfigure}
\begin{subfigure}{0.3\linewidth}
\includegraphics[width=\linewidth]{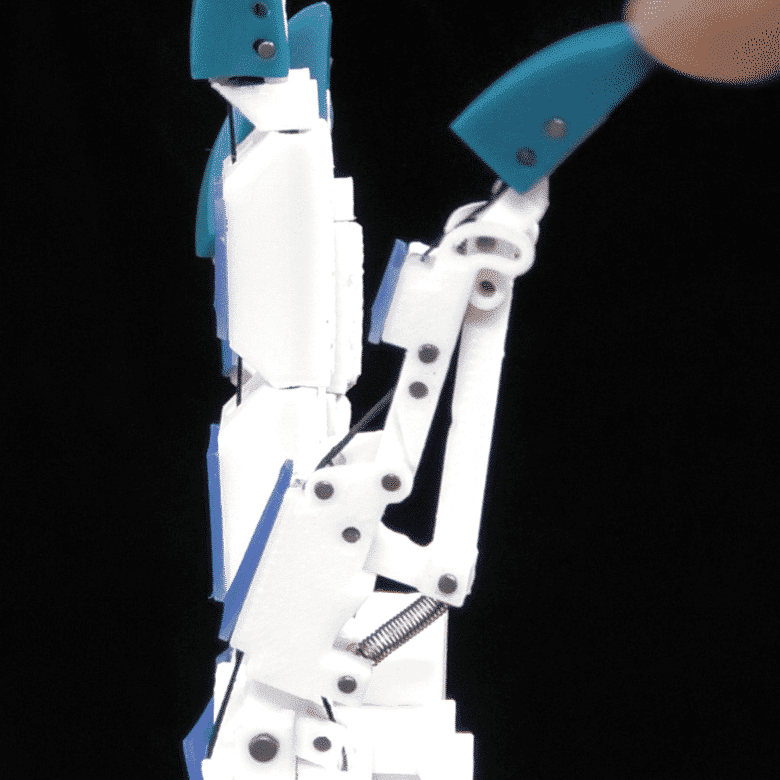}
\caption{}
\label{fig:ee}
\end{subfigure}
\begin{subfigure}{0.3\linewidth}
\includegraphics[width=\linewidth]{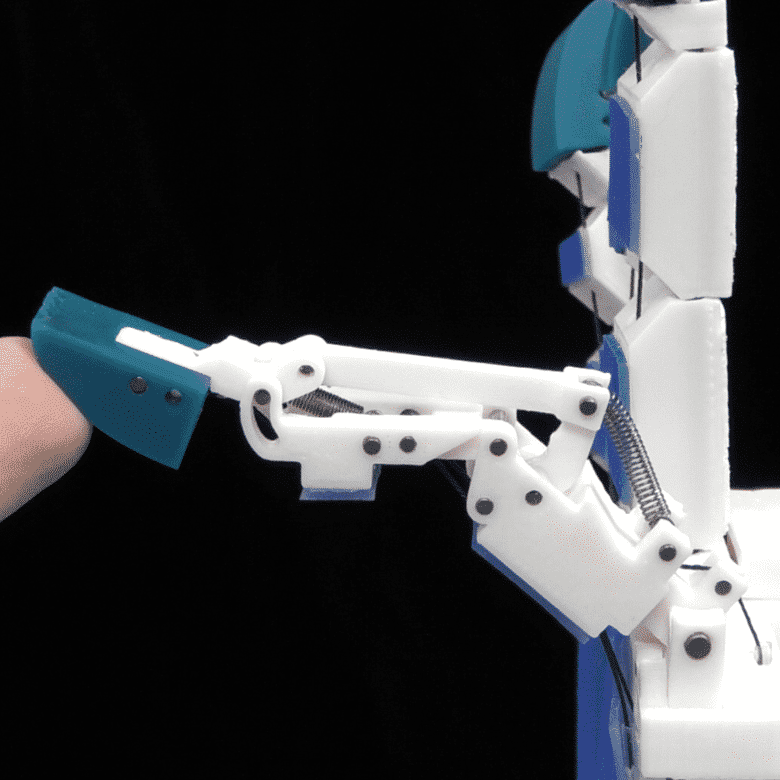}
\caption{}
\label{fig:ff}
\end{subfigure}
\caption{Illustration of different robotic hand postures. (a)-(d): robotic finger postures without the influence of external forces. (e)-(f): robotic finger postures forced by external forces on the palmar side of fingertip.}
\label{fig:robotichandposture}
\end{figure}

\begin{figure}
\centering
\begin{subfigure}[b]{0.24\linewidth}
\includegraphics[width=0.98\linewidth]{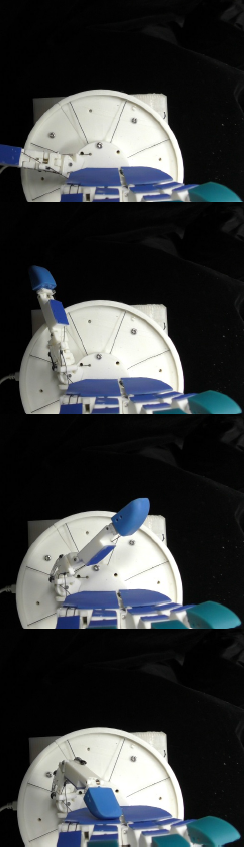}
\caption{}
\label{fig:11}
\end{subfigure}
\begin{subfigure}[b]{0.24\linewidth}
\includegraphics[width=0.98\linewidth]{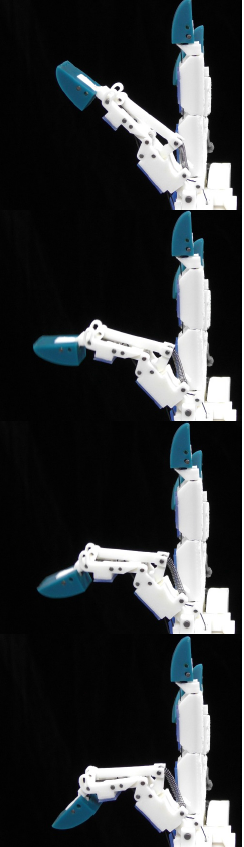}
\caption{}
\label{fig:22}
\end{subfigure}
\begin{subfigure}[b]{0.24\linewidth}
\includegraphics[width=0.98\linewidth]{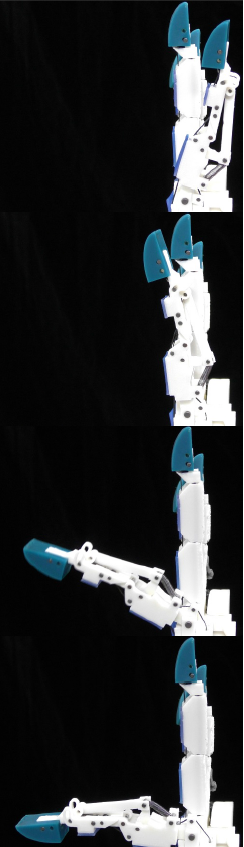}
\caption{}
\label{fig:33}
\end{subfigure}
\begin{subfigure}[b]{0.24\linewidth}
\includegraphics[width=0.98\linewidth]{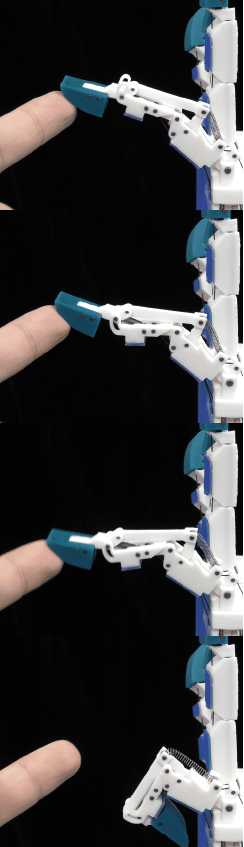}
\caption{}
\label{fig:44}
\end{subfigure}
\caption{Illustration of the main and the special movements of the index fingerand the thumb. (a): from top to bottom shows the thumb movements about the CMC joint's adduction, abduction, and flexion, as well as the MCP joint's pronation along with the MCP joint's flexion and IP joint's flexion. (b): The independent movement of IP joints. (c): the independent movement of MCP joints. (d): the process of the fingertip against external forces } 
\label{fig:fingermovement}
\end{figure}

\begin{figure*}[tb]
\centering
  \includegraphics[width=\linewidth]{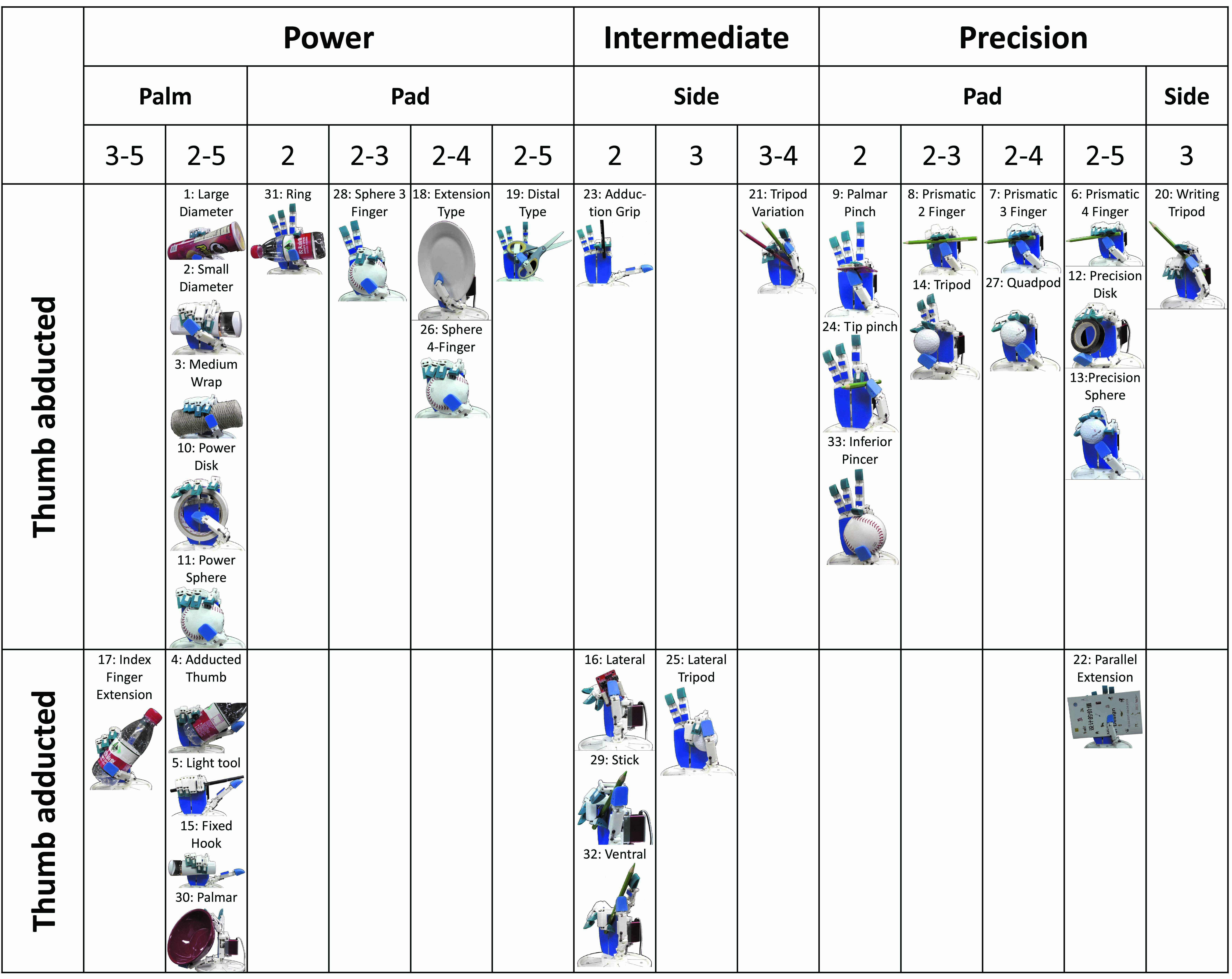}
  \caption{33 stable grasping postures achieved using our robotic hand.} 
  \label{fig:static_postures}
\end{figure*}

To evaluate the efficacy of our task-oriented deign, we first quantitatively test the index finger and thumb's movement ranges, then construct mapping relationship between robotic index finger postures and mechanism activation within the sagittal plane. Third, we show the main and special movements including thumb's pronation, index finger's independent movement of IP and MCP joints, and the compliant movement of the fingertip under external forces. Finally, we qualitatively conduct the grasping experiments to prove that our robotic hand is capable of performing 33 static and stable postures.
% The test results are reported in following subsections.

\subsection{Movement range}
The thumb initial position is shown in \prettyref{fig:whole_hand}. The ulnar side of the thumb is in the same plane as the hand's palm. The CMC, MCP, IP joints are fixed by springs at 30 degree extension, 0 degree extension and 0 degree extension respectively. We define this initial position as the position of CMC joint's 45 degree adduction. In this position, the angle difference between the proximal phalanx of the thumb and middle finger is 75 degrees in the palm plane. The degree ranges for the index and thumb are summarized in \prettyref{tab:finger_range}.

\begin{table}[!htbp]
\caption{The index and thumb's movement range}
\label{tab:finger_range}
\begin{center}
% \begin{tabular}{|c|c|c|c|}
% Finger                 & Joint                & Minimum      & Maximum      \\ \hline
% \multirow{5}{*}{Thumb} & \multirow{2}{*}{CMC} & 30 extension & 45 flexion   \\
%                        &                      & 45 abduction & 45 adduction \\
%                        & \multirow{2}{*}{MCP} & 0 extension  & 90 flexion   \\
%                        &                      & 0 supination & 45 pronation \\
%                        & IP                   & 0 extension  & 90 flexion   \\ \hline
% \multirow{4}{*}{Index} & \multirow{2}{*}{MCP} & 30 extension & 90 flexion   \\
%                        &                      & 30 abduction & 30 adduction \\
%                        & PIP                  & 0 extension  & 90 flexion   \\
%                        & DIP                  & 30 extension & 90 flexion  
% \end{tabular}
\begin{tabular}{|c|c|c|c|}
\hline
Finger                 & Joint                & Minimum      & Maximum      \\ \hline
\multirow{5}{*}{Thumb} & \multirow{2}{*}{CMC} & 30 extension & 45 flexion   \\ \cline{3-4} 
                       &                      & 45 abduction & 45 adduction \\ \cline{2-4} 
                       & \multirow{2}{*}{MCP} & 0 extension  & 90 flexion   \\ \cline{3-4} 
                       &                      & 0 supination & 45 pronation \\ \cline{2-4} 
                       & IP                   & 0 extension  & 90 flexion   \\ \hline
\multirow{4}{*}{Index} & \multirow{2}{*}{MCP} & 30 extension & 90 flexion   \\ \cline{3-4} 
                       &                      & 30 abduction & 30 adduction \\ \cline{2-4} 
                       & PIP                  & 0 extension  & 90 flexion   \\ \cline{2-4} 
                       & DIP                  & 30 extension & 90 flexion   \\ \hline
\end{tabular}
\end{center}
\end{table}

\subsection{Mapping relationship}
The cable layout is detailed in \prettyref{fig:index_whole_structure}, where the orange line mimics the human's FDP, the blue line mimics the human's FDS, and the pink line mimics the human's EDC. In sagittal plane, the robotic finger postures without external forces can be divided into 4 classes as shown in \prettyref{fig:robotichandposture}: (a) represents MCP joint's hyperextension or extension and IP joints' extension. (b) represents MCP joint's hyperextension or extension and IP joints' flexion. (c) represents MCP joint's flexion and IP joints' extension. (d) represents MCP joints' flexion and IP joints' flexion. When external force is exerted on the palmar side of the fingertip and DIP joint stays in the hyperextension state, the corresponding robotic finger postures can be divided into 2 classes as shown in \prettyref{fig:robotichandposture}: (e) represents PIP joint's extension no matter which MCP joint angle is. (f) represents PIP joint's flexion no matter which MCP joint angle is. Their mapping relationship between postures within sagittal plane and the mechanism activations is summarized in~\prettyref{tab:robotic_strategies}. However, actuators cannot provide musculotendinous tension like human hand. Thus, when the index finger flexes, the actuator which controls the finger extension must actively loose the line, but this is not illustrated in~\prettyref{tab:robotic_strategies}.
In~\prettyref{tab:robotic_strategies}, BL, OL, SP and PL represent the blue line, orange line, Spring 1 and pink link as shown in \prettyref{fig:index_whole_structure} respectively. $+$ represents the actuator actively tighten the corresponding line. BL's $(+)$ means that the actuator can actively tighten this line to resist large external forces. PL's $(+)$ means that the actuator can actively tighten this line to stabilize the MCP joint at any desired degree. SP is a purely passive structure implementing the function of PI/DI in the sagittal plane and its $(+)$ means that it provides the passive resistance in any finger posture. We can easily observe that the robotic hand's control strategies are the same as that of humans within the sagittal plane for the three tendons including FDP, FDS and EDC. Meanwhile, we use the spring to approximate PI/DI functions in flexion/extension and use servo to directly control MCP joint's abduction/adduction. In this way, we can accomplish the index finger's dexterity with only 4 actuators, and this design can also respond to external forces on the fingertip similar to humans.

\begin{table*}[!htbp]
\caption{Robotic hand's control strategies}
\label{tab:robotic_strategies}
\begin{center}
% \begin{tabular}{lcccc}
% \toprule
% Without external forces                  & BL & OL & SP & 
% PL \\
% \midrule
% MCP (hyper)extension and IP extension  &             &           & (+)           & +         \\
% MCP (hyper)extension and IP flexion    & +           & +         & (+)           & +         \\
% MCP flexion and IP extension    & +           & +         & (+)           & (+)       \\
% MCP flexion and IP flexion      & +           & +         & (+)           & (+)       \\
% \toprule
% Under external forces                 & FDP & FDS & PI/DI & EDC \\
% \midrule
% PIP extension and any MCP angle & (+)         & +         & (+)           &           \\
% PIP flexion and any MCP angle   & (+)         & +         & (+)           &          \\
% \bottomrule
% \end{tabular}
\begin{tabular}{|c|l|l|c|c|c|c|}
\hline
\multirow{2}{*}{\textbf{Posture}} & \multicolumn{2}{c|}{\multirow{2}{*}{\textbf{Description}}}                                       & \multicolumn{4}{c|}{\textbf{Mechanism}}                                     \\ \cline{4-7} 
                                  & \multicolumn{2}{c|}{}                                                                            & BL                    & OL                    & SP  & PL                    \\ \hline
\prettyref{fig:aa}                         & \multirow{4}{*}{without external forces} & MCP joint's (hyper)extension and IP joints' extension & \multicolumn{1}{l|}{} & \multicolumn{1}{l|}{} & (+) & +                     \\ \cline{1-1} \cline{3-7} 
\prettyref{fig:bb}                         &                                          & MCP joint's (hyper)extension and IP joints' flexion   & +                     & +                     & (+) & +                     \\ \cline{1-1} \cline{3-7} 
\prettyref{fig:cc}                         &                                          & MCP joint's flexion and IP joints' extension          & +                     & +                     & (+) & (+)                   \\ \cline{1-1} \cline{3-7} 
\prettyref{fig:dd}                         &                                          & MCP joint's flexion and IP joints' flexion            & +                     & +                     & (+) & (+)                   \\ \hline
\prettyref{fig:ee}                       & \multirow{2}{*}{under external forces}   & PIP joint's extension and any MCP joint's angle       & (+)                   & +                     & (+) & \multicolumn{1}{l|}{} \\ \cline{1-1} \cline{3-7} 
\prettyref{fig:ff}                       &                                          & PIP joint's flexion and any MCP joint's angle         & (+)                   & +                     & (+) & \multicolumn{1}{l|}{} \\ \hline
\end{tabular}
\end{center}
\end{table*}

\subsection{Main and special movements}
As shown in \prettyref{fig:11}, the thumb accomplishes the CMC joint's adduction, abduction, flexion, as well as MCP joint's pronation along with MCP joint's flexion and IP joint's flexion. In \prettyref{fig:22} and \prettyref{fig:33}, our robotic hand achieves the independent movements of IP and MCP joints. In addition, as shown in \prettyref{fig:44}, our index finger's fingertip can arrive at a compliant and adaptive posture under external forces and actively resist external forces by tightening the blue line which is illustrated in \prettyref{fig:index_whole_structure}.
% As shown in \prettyref{fig:whole_handbone}, when BL and OL contract, the robotic fingertip touches the human finger. If BL stops contraction and OL further contracts, the robotic fingertip state will be like \prettyref{fig:index_tendon}. If the human hand moves downwards facing the robotic fingertip's pressure and OL contracts further, the robotic fingertip state will be like \prettyref{fig:extensor_mechanism_flex}. And finally when the human finger is removed or can't resist the robotic fingertip's pressure with the help of BL, the robotic fingertip state will be like \prettyref{fig:gliding_mechanism}.

\subsection{Grasping experiments}
% Understanding the way humans grasp objects, knowing the kinematic implications and limitations associated with each grasp, and knowing common use patterns is important in many domains ranging from medicine and rehabilitation, psychology, and product design, among many others. As detailed in our motivation, this version of our hand is to solve the problem how to grasp any object which can be grasped by human single hand at least in the robotic hand's structure. Thomas Feix has found and arranged 33 different grasp types into the GRASP taxonomy which considered only static and stable grasps performed by one hand~\cite{Feix:2016:GTH}. Our robotic hand's performance is shown in \prettyref{fig:static_postures}.
One of our contributions is that the proposed robotic hand is of great dexterity and can pass all challenging tests defined in the GRASP taxonomy~\cite{Feix:2016:GTH}, where a robotic hand need to accomplish 33 standard grasping postures in different tasks. As shown in \prettyref{fig:static_postures}, our CATCH hand can successfully accomplish all the 33 static and stable grasping postures. For more details, please refer to the video.

\section{Conclusion and Future Works}
\label{sec:conclusion}

In this work, we have designed an anthropomorphic robotic hand that closely mimics the control strategies of human hand's index finger with a novel four-bar linkage to achieve independent control of IP and MCP joints. What's more, we reconstruct muscles' functions in the thumb and replace 9 human muscle tendons with 3 cables using task-oriented design to pass all challenging tests defined in the GRASP taxonomy~\cite{Feix:2016:GTH}. However, because the cable-driven design cannot decouple joints' movements, we need to calibrate the relationship between joint angles and corresponding cables' lengths, which is troublesome and time-consuming and further prevents us from achieving precise and high-speed control like motor-driven anthropomorphic hand\footnotemark.\footnotetext{\textbf{UT/HDS Hand: } https://www.youtube.com/watch?v=vu6APoC0IOA} 
%In addition, cables cannot bear large forces and are easily broken by mutual antagonism between cables which mimic flexor and extensor respectively. Meanwhile, without modular design, our hand is difficult to be repaired once the cables are broken.

In future work, we are planning to change the phalanges and the palm's design from 3D printing of hard materials to soft materials, in order to improve the hand's compliance with contact objects and to increase the grasping capability in terms of accomplishing more general grasping tasks like grasping all the YCB objects~\cite{calli2015ycb}. 
%Next, we will incorporate biomimetic wrist design and fingertip tactile sensor like Biotac~\cite{fishel2013syntouch} into our robotic hand so that we can perform manipulation studies such as how to pinch a sheet of paper from the table, which requires index fingertip's passive hyperextension DOF and tactile feedbacks. What's more, we will integrate proprioceptive sensors into our hand to learn the relationship between joint angles and corresponding cables' lengths by using artificial neural network in order to accomplish the independent precise control of every joint.

{\small
\bibliographystyle{IEEEtran}
\bibliography{reference}
}

\end{document}